\documentclass{article}
\PassOptionsToPackage{table}{xcolor}
\usepackage{iclr2027_conference,times}
\iclrfinalcopy
\pdftrailerid{}
\usepackage[T1]{fontenc}
\usepackage[utf8]{inputenc}
\usepackage{amsmath,amssymb,amsfonts,booktabs,multirow,array,longtable,tabularx}
\usepackage[table]{xcolor}
\usepackage{graphicx,microtype,hyperref,url,flafter,placeins,float,wrapfig,etoolbox}
\AtBeginEnvironment{table}{\setlength{\abovecaptionskip}{0pt}\setlength{\belowcaptionskip}{5pt}}
\definecolor{ourblue}{rgb}{0.368,0.507,0.71}
\hypersetup{colorlinks, linkcolor=ourblue, citecolor=ourblue, urlcolor=ourblue}
\newcommand{\tblhead}[1]{\textbf{\boldmath #1}}
\definecolor{oursrow}{HTML}{E5EFF9}
\definecolor{groupbg}{HTML}{FFF2CC}
\newcommand{\tblindent}[1]{\hspace{0.6em}#1}
\newcommand{\sd}[1]{{\scriptsize$\,\pm\,$#1}}
\newcommand{\kap}{\kappa}

\title{Distilling Directional Verification\vspace{14pt}}
\author{%
  \vspace*{4pt}
  \makebox[5.28in][c]{%
  \begin{tabular}{@{}c@{\hspace{0.02in}}c@{\hspace{0.02in}}c@{}}
    \parbox[c]{1.76in}{\centering Jungseob Lee$^{1}$\\[1pt] \fontsize{8pt}{7.5pt}\texttt{omanma1928@korea.ac.kr}} &\vspace*{4pt}
    \parbox[c]{1.76in}{\centering Sugyeong Eo$^{2}$\\[1pt] \fontsize{8pt}{7.5pt}\texttt{s.eo@yonsei.ac.kr}} &\vspace*{4pt}
    \parbox[c]{1.76in}{\centering Seongtae Hong$^{1}$\\[1pt] \fontsize{8pt}{7.5pt}\texttt{ghdchlwls123@korea.ac.kr}}\\[4pt]\vspace*{4pt}
    \makebox[0pt][l]{\parbox[c]{1.76in}{\centering Seungyoon Lee$^{1}$\\[1pt] \fontsize{8pt}{7.5pt}\texttt{dltmddbs100@korea.ac.kr}}} &\vspace*{4pt}
    \makebox[0pt][l]{\parbox[c]{1.76in}{\centering Chanjun Park$^{3}$\\[1pt] \fontsize{8pt}{7.5pt}\texttt{chanjun.park@ssu.ac.kr}}} &\vspace*{4pt}
    \makebox[0pt][l]{}\\[4pt]\vspace*{4pt}
    \makebox[0pt][l]{\parbox[c]{1.76in}{\centering Jaehyung Seo$^{4,\dagger}$\\[1pt] \fontsize{8pt}{7.5pt}\texttt{seojae777@konkuk.ac.kr}}} &
    \makebox[0pt][l]{\parbox[c]{1.76in}{\centering Heuiseok Lim$^{1,\dagger}$\\[1pt] \fontsize{8pt}{7.5pt}\texttt{limhseok@korea.ac.kr}}} &
    \makebox[0pt][l]{}
  \end{tabular}%
  }\\[7pt]
  \small
  \makebox[5.28in][c]{%
    \begin{tabular}{c}
      $^{1}$Korea University \hspace{0.1in} $^{2}$Yonsei University Mirae Campus\\[1pt]
      $^{3}$Soongsil University \hspace{0.1in} $^{4}$Konkuk University
    \end{tabular}%
  }%
}
\begin{document}
\maketitle
  \begingroup
  \renewcommand{\thefootnote}{\textdagger}
  \footnotetext{Corresponding authors.}
  \endgroup
  \lhead{Preprint}

\begin{abstract}
Knowledge distillation aims to transfer the factual knowledge of large language models to smaller models for efficient deployment. Yet a teacher may recall a relation in one direction while failing to generate the answer in the reverse direction. Distillation from its generated answers can therefore propagate this directional limitation to the student. The same teacher can nevertheless recognize such an answer by scoring the relation in the direction it knows. We introduce directional label distillation, in which frozen teachers score candidate answers in that known direction and the best-scoring candidate becomes the student's training target. On facts about parents and their children, known-direction scoring yields more accurate labels than scoring the requested direction, even after tuned corrections for name priors. With prior-corrected scores, the better direction depends on the facts rather than the template, and reverses on mined facts whose notable entity is the parent rather than the child. With the evaluated children's forward facts withheld, students trained on known-direction labels improve open-ended accuracy on their trained queries by 13 to 15 points over students trained on prior-corrected reverse labels. After generated answers are matched to a fixed name list by lexical similarity, students reproduce nearly all selected labels. Their accuracy largely follows label quality. The label advantage holds on unscreened queries and when candidates are retrieved without inserting correct answers. Our findings show that directional verification mitigates the transfer of errors from teacher-generated answers to students by providing more accurate training targets.
Code is available at \url{https://github.com/js-lee-AI/directional-verification}.
\end{abstract}

\newcommand{\figoverview}{%
\begin{figure}[t]
\centering
\setlength{\abovecaptionskip}{6pt}
\includegraphics[width=\linewidth,trim=0 12bp 0 0,clip]{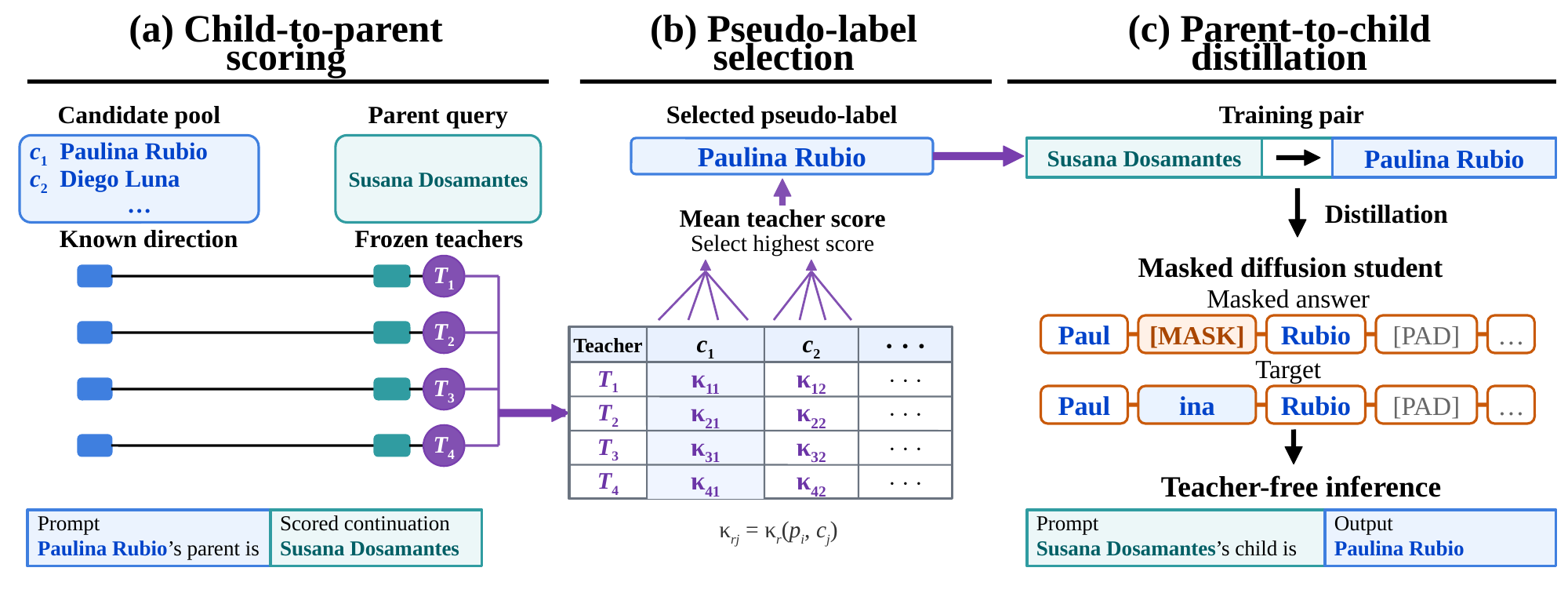}\par
\caption{Directional label distillation with four teacher channels. Frozen teachers score the same query parent after each candidate child (a). The candidate with the highest mean teacher score becomes the pseudo-label (b). A masked diffusion student learns to reconstruct masked tokens of the selected answer and answers reverse queries without teachers at inference (c). Blue and teal mark child and parent names.}
\label{fig:example}
\end{figure}
}
\figoverview

\section{Introduction}
\label{sec:intro}
Large language models acquire knowledge about relations between entities, yet learning a relation in one direction does not ensure that they can use it in reverse \citep{berglund2023reversal,wang2025reversalbinding}. This reversal curse limits their ability to answer questions using facts they have already learned \citep{berglund2023reversal,kitouni2024factorization,wang2025reversalbinding}. Distillation trains students to match teacher predictions \citep{kim2016sequence,agarwal2023gkd}. Using incorrect reverse answers as training targets can pass errors to the student even when the teacher recalls the same facts in the other direction. Prior work links reversal failures to how models are trained to predict tokens \citep{kitouni2024factorization,zhu2024trainingdynamics}. In standard next-token training, each token is predicted from the preceding tokens \citep{kitouni2024factorization,ma2026breaking}. The same sequence does not directly train the model to predict earlier tokens from later context. Proposed remedies therefore change the training data, for example by adding reversed or identity-bridging examples \citep{golovneva2024reverse,ma2026breaking}, or change the objective, for example by predicting tokens in any order as masked diffusion models do \citep{kitouni2024factorization,nie2025llada,gong2024scaling}. Both routes change how a model learns from training data. Neither addresses how to obtain accurate reverse labels from a pretrained teacher whose reverse answers are unreliable.

A complementary route is to supply reverse supervision through knowledge distillation. For example, sequence-level and on-policy distillation train students using sequences generated by the teacher or the student \citep{kim2016sequence,agarwal2023gkd}. Best-of-$N$ distillation and reward-filtered self-training favor higher-scoring generations using reward or verification signals \citep{sessa2024bond,dong2023raft,gulcehre2023rest,singh2023restem}. However, learning from generated sequences does not by itself ensure accurate supervision for reverse queries that the teacher struggles to answer. Filtering or reranking also cannot recover a correct answer that is absent from the generated candidates \citep{cobbe2021training,lightman2023lets}.

\setlength{\textfloatsep}{12pt plus 2pt minus 2pt}

A teacher that cannot generate an answer may nevertheless recognize it. Generation and verification can disagree within a single language model \citep{rankalign2025,lee2026dart_routing}. We observe this disagreement in directional facts, where teachers struggle to name a child from its parent yet can recognize the relation in the child-to-parent direction. Building on this observation, we leverage verification instead of generation to obtain reverse targets. Candidate answers come from a fixed inventory, and the teacher scores each candidate in the direction it knows. Such a score has the form of a noisy-channel score, which rates the query under each candidate and is less sensitive to candidate priors than direct scoring \citep{min2022noisy}. Direct scores can instead be corrected for these priors \citep{holtzman2021surface,zhao2021calibrate}. When both directional probabilities and the name prior come from the same joint distribution, the two approaches rank candidates identically. However, in our experiments, pretrained teachers rank the same candidates differently across the two sentence directions, even after name-prior correction. We therefore compare the accuracy of the labels selected in each direction and test whether training on the more accurate labels improves student accuracy on reverse queries.

We introduce directional label distillation, illustrated in Figure~\ref{fig:example}. For each parent query, frozen teachers retrieve a small set of candidate children from a name inventory and score the query parent as the continuation of \texttt{\{child\}'s parent is} for every candidate. We call this child-to-parent scoring the \emph{known direction}. Averaged teacher scores select one candidate as a pseudo-label, and a student learns the resulting parent-to-child pairs without access to the teachers at inference. We use masked diffusion students because any-order training has been proposed to mitigate reversal failures \citep{kitouni2024factorization}. This choice lets us test whether more accurate reverse labels improve answer accuracy even under that objective. For comparison, we train an autoregressive student on the same label sets. To isolate the effect of label quality, our main student comparisons exclude child-to-parent training examples for the evaluated relations.

We evaluate our method on screened and unscreened cohorts of parent--child facts from Wikidata \citep{vrandecic2014wikidata}. Holding the candidate pool and the selection rule fixed, we find that the scoring direction matters considerably more than how teacher scores are combined. Known-direction labels are more accurate than reverse labels under every aggregation rule. They also outperform labels from reverse scores with tuned prior correction, teacher generation in either direction, and teacherless name heuristics.

Our analysis of name priors helps explain this advantage. Uncorrected reverse scores tend to favor names that are likely even without the queried parent. Known-direction scoring instead evaluates the same parent name after each candidate child. This avoids directly rewarding candidates for the probability of their own names. The advantage also holds on an unscreened cohort whose candidate lists favor neither direction, and when teachers retrieve candidates from the full inventory without an inserted answer. Our additional analyses using prior-corrected scores show that the better sentence direction remains unchanged when the query is reversed, but switches between facts about notable children and facts about notable parents. These findings motivate choosing the verification direction according to the teacher's knowledge of the queried fact.

We train students to answer parent-to-child queries under an identical recipe. Their labels are selected using either known-direction (child-to-parent) teacher scores or prior-corrected reverse scores. On trained queries in both cohorts, known-direction labels yield 13 to 15 points higher open-ended accuracy than prior-corrected reverse labels. The gains persist when the entire generated answer must equal a true name. After matching generated answers to inventory names by lexical similarity, we find that students reproduce nearly all selected labels. Student accuracy largely follows label quality. In separate controls, students trained without parent-to-child answer labels remain below $3\%$ open-ended accuracy, including with any-order masked training. Directional verification thus enables a teacher to supply substantial supervision for reverse queries that its own completions rarely answer. Beyond this parent--child setting, our findings point to a broader approach to the reversal curse by using verification to turn existing relational knowledge into reverse supervision.

\setlength{\textfloatsep}{20pt plus 2pt minus 4pt}
\section{Directional label distillation}
\label{sec:method}
Given a parent query $p_i$, the student must name a true child. Frozen teachers select a target $\hat c_i$ from a fixed name inventory $\mathcal I$, without using the recorded answer for selection. We score each candidate child in the child-to-parent direction, then train the student to answer in the parent-to-child direction. Teacher $T_r$ evaluates the query parent's probability after \texttt{\{child\}'s parent is} for each candidate child $c$. The known-direction score averages the parent's token log probabilities,
\begin{equation}
\kap_r(p_i,c)=\frac{1}{|\operatorname{tok}_r(p_i)|}
\log P_{T_r}\!\left(p_i\mid t_{\mathrm{parent}}(c)\right),
\label{eq:score}
\end{equation}
where $t_{\mathrm{parent}}(c)$ denotes this prompt and $|\operatorname{tok}_r(p_i)|$ is the number of parent tokens. The parent continuation stays fixed across candidates, while the child context changes. Each teacher--template pair defines a scoring channel.

\subsection{Candidate acquisition and label construction}
\label{sec:aggregation}
We first form a candidate pool and then select a training label by combining teacher scores. For the screened cohort, Qwen3-8B \citep{qwen2025qwen3}, OLMo-2-7B \citep{olmo2024olmo2}, and Mistral-7B-v0.3 \citep{jiang2023mistral} independently scan the full inventory using known-direction scores. The union of each teacher's eight highest-scoring candidates forms $\mathcal C_i$. Llama-3.1-8B-Instruct \citep{llama2024herd} scores this pool without adding candidates.

We average the four main-template scores and select the highest-scoring candidate other than the queried parent as the training label,
\begin{equation}
\hat c_i=\mathop{\arg\max}_{c\in\mathcal C_i,\ c\ne p_i}
\frac{1}{4}\sum_{r=1}^{4}\kap_r(p_i,c).
\label{eq:primarylabel}
\end{equation}
Figure~\ref{fig:example} illustrates the full pipeline with this four-channel mean for label selection. Our primary unscreened comparison uses pools fixed before teacher scoring. In matched comparisons, we hold the candidate pool and selection rule fixed, retaining queries whose pool contains no correct answer.

We compare six aggregation rules on the screened cohort to separate the effects of aggregation and scoring direction. Using the same four main-template channels, we evaluate raw-score, z-score, and probability means, Borda, reciprocal-rank fusion (RRF), and soft agreement. Soft agreement estimates channel weights from how consistently the channels support the same candidates, without using gold answers. An eight-channel variant uses two templates for each teacher and soft agreement to supply labels for student controls that retain all forward training facts. The appendix gives the prompt templates and aggregation formulas.

\subsection{Reverse comparators}
\label{sec:reverse}
Reverse scores evaluate the requested continuation. Writing $\ell_r(y\mid t)$ for the mean continuation-token log probability of $y$ after prompt $t$, the reverse score of a candidate is $\ell_r(c\mid t_{\mathrm{child}}(p_i))$, where $t_{\mathrm{child}}(p_i)$ is \texttt{\{parent\}'s child is}. This score also rewards names that are probable on their own, so we subtract a candidate's context-only score, following domain-conditional pointwise mutual information scoring \citep{holtzman2021surface}:
\begin{equation}
s_r^{\lambda}(p_i,c)=\ell_r(c\mid t_{\mathrm{child}}(p_i))-\lambda\,\ell_r(c\mid\texttt{The child is}).
\label{eq:dc}
\end{equation}
The domain-context (DC) reverse score sets $\lambda=1$. Summed-token versions replace the means with sums, because candidate children differ in token length whereas the parent continuation of the known direction is fixed within a query. Tuned reverse scores choose $\lambda$ from a grid and may replace the context-only term with a Monte Carlo estimate of the candidate's marginal log probability after unrelated parents; their form and $\lambda$ are tuned with gold labels on other query sets and then transferred. Teacher generation supplies a further comparator, in which each teacher completes the raw prompt \texttt{\{parent\}'s child is}, greedily and by sampling, and the completion is mapped onto the candidate list.

\subsection{Student training and inference}
Reverse training pairs $(p_i,\hat c_i)$ supervise a student as completions of \texttt{\{parent\}'s child is}, while forward examples complete \texttt{\{child\}'s parent is} with a parent name; in both, the prompt stays visible and only the answer name is scored. Because any-order masked training has been proposed as a remedy for reversal failures \citep{kitouni2024factorization}, our main students are masked diffusion models (MDMs) initialized from Qwen3-0.6B or Qwen3-4B, following the autoregressive-to-diffusion adaptation of \citet{gong2024scaling}. Each adds a mask token, employs bidirectional attention, and predicts masked answer slots with cross-entropy \citep{austin2021structured,sahoo2024simple}. The answer occupies ten slots, of which a uniformly sampled number from one to ten is masked and scored. A forward warm stage teaches the format, and true forward examples remain mixed into reverse supervised fine-tuning (SFT).

At inference the teachers are discarded, and the MDM fills one slot at a time in confidence order over ten model passes. We evaluate its open string and an \emph{inventory-matched} answer obtained by lexical postprocessing. An autoregressive (AR) student trained on the same labels provides a second prediction objective, with the separate token budget given in Appendix~\ref{app:implementation}.

\section{Experimental setup}
\label{sec:setup}
We measure label accuracy and student internalization separately. Scoring channels and aggregation rules were selected with gold evaluations on the screened cohort, whereas the unscreened cohort and its candidate policies were fixed before any teacher inference. We call a comparison preregistered when its analysis plan was fixed before the relevant outcomes were computed.

\paragraph{Data and candidate pools.}
The corpus contains $10{,}505$ Wikidata parent--child pairs \citep{vrandecic2014wikidata}, $8{,}218$ distinct children, and $9{,}918$ parents. Children are selected for having English Wikipedia articles. The Qwen3-8B screen keeps a fact when its best true-parent score exceeds every distractor score, retaining $2{,}476$ of $4{,}400$ screened children. A fixed pseudorandom shuffle orders these facts, and its first $1{,}500$ pairs become the reverse-label training set. The acquired pool contains $28{,}463$ candidate slots and an accepted answer for $97.13\%$ of queries, and comparisons on it condition on the screen and the acquired inventory.

The unscreened cohort contains $512$ new parent queries without a forward screen. Each query has two direction-independent lists of $64$ names, with uniformly sampled or lexically similar distractors. Both lists include the recorded answer, and the cohort also trains separate students, with its construction and paired tests described in Appendix~\ref{app:prospective}.

\begin{table}[!t]
\caption{MDM-0.6B students trained on selected reverse labels (\%, three-seed mean $\pm$ sample SD). $x$ is label accuracy, and the last two columns give the share of outputs that reproduce the selected label as a whole string or after inventory matching. Blue marks the primary known-direction labels.}
\label{tab:factual}
\centering\small
\setlength{\tabcolsep}{2.6pt}
\begin{tabular*}{\linewidth}{@{\extracolsep{\fill}}lrrrrrr@{}}
\toprule
 & & \multicolumn{3}{c}{\tblhead{Accuracy against true children}} & \multicolumn{2}{c}{\tblhead{Reproduces label}}\\
\cmidrule(lr){3-5}\cmidrule(l){6-7}
\tblhead{Label source} & \tblhead{$x$} & \tblhead{Open} & \tblhead{Whole} & \tblhead{Matched} & \tblhead{String} & \tblhead{Inventory}\\
\midrule
\rowcolor{groupbg}[0pt][0pt]\multicolumn{7}{@{}c@{}}{\textit{Acquired pools}\strut}\\
Reverse, mean token & 53.60 & 47.17\sd{1.29} & 47.00\sd{1.27} & 53.14\sd{0.41} & 79.59\sd{1.74} & 97.67\sd{0.30}\\
DC reverse, mean token & 54.82 & 49.26\sd{0.08} & 49.06\sd{0.07} & 53.93\sd{0.48} & 90.36\sd{0.29} & 97.65\sd{0.79}\\
DC reverse, summed token & 75.18 & 63.26\sd{1.30} & 63.14\sd{1.37} & 73.98\sd{0.23} & 79.98\sd{1.35} & 97.39\sd{0.51}\\
\addlinespace[2pt]
\rowcolor{oursrow}[0pt][0pt]
\textbf{Known direction, mean token} & 90.72 & 78.03\sd{2.86} & 77.75\sd{2.91} & 88.94\sd{0.46} & 84.56\sd{3.02} & 97.70\sd{0.61}\\
\tblindent{Llama two-template mean} & 90.29 & 75.11\sd{3.49} & 74.87\sd{3.42} & 88.23\sd{0.76} & 81.61\sd{3.94} & 97.24\sd{0.91}\\
\tblindent{Eight-channel agreement} & 91.58 & 79.09\sd{0.80} & 78.78\sd{0.59} & 89.78\sd{0.07} & 84.96\sd{0.92} & 97.67\sd{0.18}\\
\midrule
\rowcolor{groupbg}[0pt][0pt]\multicolumn{7}{@{}c@{}}{\textit{Uniform lists}\strut}\\
DC reverse, summed token & 73.44 & 71.88\sd{0.59} & 71.88\sd{0.59} & 73.24\sd{0.00} & 97.33\sd{0.11} & 99.74\sd{0.11}\\
Tuned reverse, transferred & 74.61 & 72.20\sd{0.81} & 72.07\sd{0.78} & 74.35\sd{0.11} & 96.81\sd{1.27} & 99.61\sd{0.20}\\
\rowcolor{oursrow}[0pt][0pt]
\textbf{Known direction, mean token} & 87.89 & 85.29\sd{0.56} & 85.22\sd{0.63} & 87.50\sd{0.20} & 96.81\sd{0.60} & 99.48\sd{0.11}\\
\addlinespace[2pt]
\rowcolor{groupbg}[0pt][0pt]\multicolumn{7}{@{}c@{}}{\textit{Lexical lists}\strut}\\
DC reverse, summed token & 55.86 & 54.49\sd{0.70} & 54.49\sd{0.70} & 55.79\sd{0.11} & 97.07\sd{0.68} & 99.67\sd{0.30}\\
Tuned reverse, transferred & 57.23 & 55.66\sd{0.59} & 55.60\sd{0.49} & 57.10\sd{0.23} & 97.59\sd{0.69} & 99.87\sd{0.23}\\
\rowcolor{oursrow}[0pt][0pt]
\textbf{Known direction, mean token} & 71.29 & 69.86\sd{0.45} & 69.79\sd{0.56} & 71.09\sd{0.20} & 97.59\sd{0.69} & 99.80\sd{0.20}\\
\bottomrule
\end{tabular*}
\end{table}

\paragraph{Supervision and exposure.}
For the primary student comparison, we remove every forward edge involving any child of the first $2{,}000$ facts in that order, which include every trained child, before both training stages, leaving $7{,}888$ pairs, and train on all $1{,}500$ reverse pseudo-labels. Because retained siblings may still answer some queries, the primary evaluation uses the $1{,}390$ trained queries whose parent appears in no retained forward example, which we call exposure-free. Students on the unscreened cohort withhold the forward edges of every child of a query parent, so all $512$ trained queries per pool are exposure-free. Default-exposure control students instead train on all true forward pairs. A gold-reverse oracle serves as a reference arm, and none of these controls removes pretraining information. Student tables group their rows by three evaluation sets named here: the acquired pools on the screened cohort's $1{,}390$ exposure-free queries, and the uniform and lexical lists on the unscreened cohort's $512$ queries.

\paragraph{Evaluation.}
Open accuracy uses accent- and case-normalized full-name substring matching against any true child of the query parent, without bare-surname credit. Whole-answer accuracy requires the entire normalized output to equal a true child's name. Inventory-matched accuracy first selects a name by lexical similarity and then requires exact membership in the query parent's true-child set. The normalization rules are listed in Appendix~\ref{app:exactanswer}.

\paragraph{Optimization.}
MDMs use low-rank adaptation (LoRA) \citep{hu2022lora} with rank $64$, AdamW \citep{loshchilov2019decoupled} at $10^{-4}$, batch size $16$, $4{,}000$ warm steps, and $4{,}800$ mixed SFT steps. Student comparisons use 0.6B MDMs and three random seeds, with GPU class matched within each paired run. The mixed corpus repeats each reverse example until the reverse and forward examples are roughly balanced, five times on the screened cohort and nineteen times on the unscreened cohort. Student spreads are sample standard deviations across training seeds, and paired differences compare the same seed. Label comparisons instead resample query outcomes, clustering repeated parents in the acquired cohort.

\section{Results}
\label{sec:results}
\subsection{Students trained on directional labels}
\label{sec:internalization}
\begin{figure}[t]
\centering
\includegraphics[width=\linewidth]{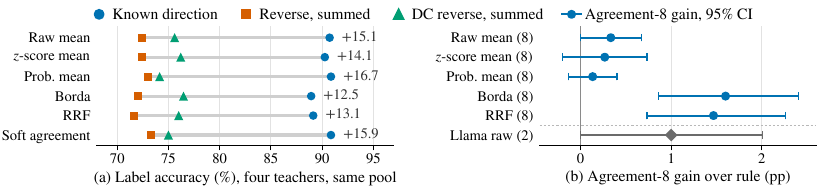}\par
\caption{Label accuracy on the $1{,}500$ acquired pools. (a) Scoring directions under six aggregation rules. (b) Eight-channel agreement gains over simpler rules on the same vectors and a two-vector Llama mean, with paired parent-cluster 95\% intervals.}
\label{fig:controls}
\end{figure}

Table~\ref{tab:factual} compares student performance across label sources within each cohort, with the training recipe held fixed. On the screened cohort's exposure-free queries, known-direction labels improve open accuracy by $14.77$ points over summed-token DC labels, the most accurate reverse labels without a tuned coefficient. On the unscreened cohort, the gains are $13.09$ to $15.36$ points over summed DC and transferred tuned-reverse labels. These gains hold in every paired seed and under whole-answer evaluation, indicating that they do not depend on extra text around a correct name.

Table~\ref{tab:factual} also reports label accuracy and how often students reproduce their selected labels. After inventory matching, every student returns its selected label for more than $97\%$ of queries on average over seeds, despite large differences in label accuracy. Student accuracy therefore follows label accuracy, and better labels account for most of the student gain. Further analysis of queries with disagreeing labels appears in Appendix~\ref{app:labelinternalization}.

\subsection{Scoring direction against aggregation rules}
\label{sec:direction}

Figure~\ref{fig:controls} compares scoring directions under six aggregation rules with the same four teachers and acquired pools, then compares eight-channel agreement with simpler rules. Known-direction labels exceed summed DC labels by $12.47$ to $16.73$ points under all six rules, whereas the rules differ by at most $2.33$ points within a score form. On the same eight score vectors, agreement performs on par with simple score means, so direction accounts for the label advantage. Because known-direction scans retrieved these pools, the comparison conditions on those candidates, a restriction removed by the unscreened lists and full-inventory scans.

\begin{figure}[t]
\centering
\includegraphics[width=\linewidth]{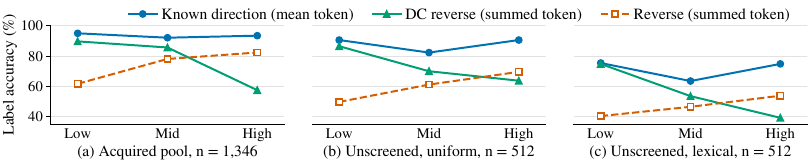}\par
\caption{Label accuracy by tercile of the true child's context-only name score. Reverse and DC reverse sum token scores, and the known direction uses mean tokens. For acquired pools, this analysis excludes queries whose true child is absent from the candidates.}
\label{fig:nameprior}
\end{figure}

\begin{figure}[t]
\centering
\includegraphics[width=\linewidth]{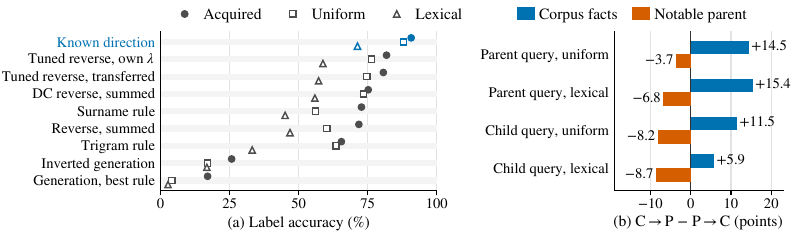}\par
\caption{Stronger comparators and sentence direction. (a) Label accuracy on the acquired pools and the uniform and lexical unscreened lists. (b) Child-to-parent sentence accuracy minus parent-to-child sentence accuracy on corpus facts and mined facts with a notable parent.}
\label{fig:comparators}
\end{figure}

\paragraph{Name priors and reverse errors.}
Figure~\ref{fig:nameprior} compares known-direction and reverse label accuracy across strata of the true child's context-only name score. A reverse score evaluates a different continuation for each candidate, mixing relational evidence with how probable a name is on its own. The stored scores for the Susana Dosamantes query illustrated in Figure~\ref{fig:example} show this mixture. Among its $17$ candidates, Selena Gomez is the most probable name on its own, above the recorded child Paulina Rubio. Scoring the requested direction selects Diego Luna, whose name is more probable than the true child's, and subtracting the prior moves the choice to Odiseo Bichir, whose name is less probable. The known direction scores the same parent continuation after every candidate and ranks the true child first. Across queries, summed reverse accuracy rises with the true child's prior, and over $90\%$ of its errors select a name with a higher prior than the true child. Subtracting the prior over-corrects, so summed DC is accurate for low-prior children but loses more than $30$ points for high-prior children on the acquired pool. Known-direction accuracy is similar in the low- and high-prior strata. Its advantage over summed DC thus grows with the prior, whereas continuation-length strata in Appendix~\ref{app:strata} show no comparable increase.

\subsection{Stronger reverse comparators}
\label{sec:comparators}

Figure~\ref{fig:comparators}a compares label accuracy from tuned reverse scores, teacher generation in either direction, and teacherless rules against the known direction on the acquired pools and both unscreened lists. Because reverse errors follow name priors, we fit the correction rather than fix it at one, using gold labels on other cohorts before transferring it. The known direction receives no such supervision, but scores every candidate, costing one teacher call for each candidate against one completion for a generated label. Tuned reverse scores improve on summed DC, yet the known direction still leads the transferred tuned scores by $10.07$ to $14.06$ points in the three cohorts, and tuning on each cohort's own gold labels leaves a similar gap. Teacher generation is a far weaker label source, because greedy and sampled completions of the raw prompt \texttt{\{parent\}'s child is} rarely name a true child, and mapping them onto the candidate list recovers few correct labels. Inverting generation, by counting candidates whose own completions of \texttt{\{child\}'s parent is} name the query parent, gives precise votes for fewer than three in ten queries. Teacherless surname and trigram rules exploit name cues yet also trail the known direction in every cohort. All comparators, their prompts, and their tuned coefficients are reported in Appendix~\ref{app:comparators}.

Figure~\ref{fig:comparators}a also compares scoring directions on the unscreened lists, which are built without teacher scores and include the recorded answer. The known direction leads every tested reverse scorer in both pools, indicating that its advantage does not depend on known-direction retrieval. Checks across all four teachers and both preregistered template pairs also favor the known direction over summed reverse and summed DC scores, as detailed in Appendix~\ref{app:prospective}.

\begingroup
\setlength{\intextsep}{2pt}
\setlength{\columnsep}{10pt}
\begin{wraptable}{r}{0.35\linewidth}
\setlength{\abovecaptionskip}{0pt}
\setlength{\belowcaptionskip}{4pt}
\caption{Full-inventory scans (\%).}
\label{tab:fullscansummary}
\centering\fontsize{8}{9}\selectfont
\setlength{\tabcolsep}{1.5pt}
\begin{tabular*}{\linewidth}{@{\extracolsep{\fill}}lrrr@{}}
\toprule
\tblhead{Score} & \tblhead{Cov.} & \tblhead{\shortstack[r]{Label\\acc.}} & \tblhead{\shortstack[r]{Union\\acc.}}\\
\midrule
\rowcolor{oursrow}[0pt][0pt]
\textbf{Known direction} & 80.66 & 66.21 & 66.02\\
DC reverse, summed & 57.03 & 39.06 & 40.43\\
\bottomrule
\end{tabular*}
\end{wraptable}
Table~\ref{tab:fullscansummary} compares full-inventory retrieval and label selection on $512$ unscreened queries without inserted answers. Coverage is the share of queries whose pool contains a true child, and union accuracy uses both pools' union. Known-direction scoring improves candidate coverage and retains a $25.59$-point advantage over summed DC when both rules use the same union pool. These results show gains in both retrieval and selection, while candidate coverage still limits accuracy. Scan details and teacher-level comparisons appear in Appendix~\ref{app:prospective}.
\par
\endgroup

\begin{table}[t]
\caption{Default-exposure student controls on all $1{,}500$ trained queries (\%, three-seed mean $\pm$ sample SD). (a) MDM-4B students trained on reverse labels with label accuracy $x$. (b) Students trained without reverse labels. Blue marks eight-channel agreement labels.}
\label{tab:historicalfactual}
\centering\small
\setlength{\tabcolsep}{2.5pt}
\begin{minipage}[t]{0.5\linewidth}
\centering
\textbf{(a) Reverse labels, MDM-4B}\par\smallskip
\begin{tabular*}{\linewidth}{@{\extracolsep{\fill}}lrrr@{}}
\toprule
\tblhead{Labels} & \tblhead{$x$} & \tblhead{Open} & \tblhead{Matched}\\
\midrule
Random & 14.13\sd{0.42} & 13.11\sd{0.47} & 13.91\sd{0.49}\\
Self-ranked & 22.60 & 21.38\sd{0.21} & 22.64\sd{0.08}\\
Single teacher & 81.13 & 75.22\sd{0.37} & 80.51\sd{0.20}\\
\rowcolor{oursrow}[0pt][0pt]
Agreement (8) & 91.53 & 84.33\sd{1.62} & 90.73\sd{0.24}\\
Gold & 100.00 & 92.00\sd{2.07} & 98.91\sd{0.04}\\
\bottomrule
\end{tabular*}
\end{minipage}\hfill
\begin{minipage}[t]{0.465\linewidth}
\centering
\textbf{(b) No reverse labels}\par\smallskip
\begin{tabular*}{\linewidth}{@{\extracolsep{\fill}}llrr@{}}
\toprule
\tblhead{Control} & \tblhead{Student} & \tblhead{Open} & \tblhead{Matched}\\
\midrule
Warm start & MDM-4B & 1.58\sd{0.14} & 39.56\sd{3.00}\\
MLM-U & MDM-4B & 1.82\sd{0.53} & 28.16\sd{6.93}\\
Warm start & AR-8B & 2.02\sd{0.14} & 37.31\sd{0.73}\\
Identity bridge & AR-8B & 1.31\sd{0.27} & 44.38\sd{1.20}\\
\bottomrule
\end{tabular*}
\end{minipage}
\end{table}

\subsection{Sentence direction and notability}
\label{sec:facts}
Figure~\ref{fig:comparators}b compares the two sentence directions for parent and child queries on corpus facts and mined facts with a notable parent. The child-to-parent (C$\to$P) sentence conditions on the child and the parent-to-child (P$\to$C) sentence on the parent. For either query side, the sentence that scores the query entity uses the channel form, and the other sentence uses the summed, context-corrected direct form. On the unscreened corpus facts, where every child has an English Wikipedia article, the C$\to$P sentence wins for parent and child queries alike. We then mined $1{,}024$ new Wikidata facts whose parent has an English article and at least $20$ sitelinks and whose child has no English article. On these facts the P$\to$C sentence wins on both lists for both query sides, and the mean difference across facts shifts by $18.68$ points toward P$\to$C. With prior-corrected direct scores, the better sentence is therefore the same for both query sides within a fact set, and it switches between the two fact sets, whose notable entities differ. The two fact sets share the template pair, list construction, and scoring forms. Era and name distribution also differ, so the comparison identifies the switch between fact sets rather than a single controlled cause. One explanation is that pretraining text about a notable person more often makes that person the subject of a sentence, giving the teachers more evidence for the sentence that begins with the notable entity.

\section{Supporting controls}
\label{sec:analysis}
\label{sec:fidelity}
\subsection{Label quality and surface realization}

Table~\ref{tab:historicalfactual} compares label sources and controls without reverse labels under default exposure, which retains every true forward pair. In panel a, matched accuracy stays within about one point of label accuracy from random to gold labels. In panel b, format adaptation, masked language modeling with a uniformly sampled mask count (MLM-U), and identity-bridge data all leave open accuracy below $3\%$. Reverse-label supervision drives the improvement in these controls. MLM-U directly tests the proposed any-order remedy for reversal failures, yet leaves these reverse queries largely unanswered. Additional controls retain the single-teacher, agreement, and gold ordering across student sizes and objectives, while larger MDM students mainly improve the expression of correct names. Stronger teacher pipelines mainly improve label accuracy, and the student's own base model can also supply useful known-direction labels. Longer training brings matched student accuracy closer to label accuracy. These comparisons are detailed in Appendix~\ref{app:controls}.

\subsection{Name cues and decoding}
\begin{figure}[t]
\centering
\includegraphics[width=\linewidth]{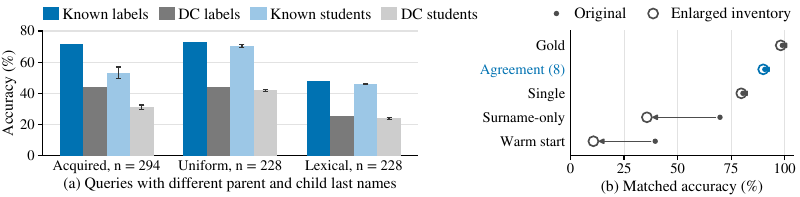}\par
\caption{Name-cue controls. (a) Label accuracy and MDM-0.6B student open accuracy on queries whose parent and recorded child have different last names. (b) Default-exposure MDM-4B matched accuracy before and after parent-only names enlarge the matching inventory.}
\label{fig:namecues}
\end{figure}

Figure~\ref{fig:namecues} tests surname cues with surname-mismatched queries in panel a and an enlarged matching inventory in panel b. On queries whose parent and recorded child have different last names, known-direction labels remain far more accurate than summed-DC labels in all three cohorts, and students trained on them outperform summed-DC students by $22$ to $29$ points on average, with a gain in every seed. Inventory matching could also exploit surname cues, so we enlarge the matching inventory with parent-only names. The enlarged inventory halves the accuracy of surname-only lookup and cuts warm-start accuracy by roughly three quarters, yet it changes trained-student accuracy by less than one point. In separate controls, decoding order affects open accuracy more than inventory-matched accuracy, as detailed in Appendix~\ref{app:controls}.

\section{Related work}
\label{sec:related}
\paragraph{Directional access and adaptation.}
Reversal studies analyze failures to infer a reverse relation after learning its forward form \citep{berglund2023reversal,wang2025reversalbinding}, and factorization and next-token prediction provide broader accounts of such failures \citep{kitouni2024factorization,bachmann2024pitfalls}. Reverse training and identity-bridge data alter the examples used for adaptation \citep{golovneva2024reverse,ma2026breaking}, while diffusion models and AR-to-diffusion adaptation offer alternative prediction orders \citep{nie2025llada,gong2024scaling}. We instead use frozen directional scores to construct the reverse targets and test them without the evaluated children's forward edges. Our controls add a measurement to this line: an any-order objective alone leaves reverse accuracy near zero on our facts, whereas the same student answers once reverse labels exist.

\paragraph{Verification and weak supervision.}
Verifiers select answers from candidate generations \citep{cobbe2021training,lightman2023lets}, and generators and validators of the same model can disagree \citep{rankalign2025}. Our candidates instead come from a name inventory, and the score evaluates the relation in the opposite direction from the requested answer. This has the form of noisy-channel scoring, which rates the input given each label and resists label priors better than direct scoring in few-shot classification \citep{min2022noisy}. Domain-conditional scoring and contextual calibration instead remove an estimated prior from direct predictions \citep{holtzman2021surface,zhao2021calibrate}, and they motivate our corrected reverse comparators. Holding candidate identities fixed lets us measure how far pretrained teachers depart from the equivalence of channel scores and prior-corrected direct scores under a coherent joint distribution. Weak supervision combines noisy evidence sources \citep{dawid1979maximum,ratner2016data}, and \citet{weaver2025shrinking} and \citet{lee2026fuse} apply related ideas to verifier ensembles, with Weaver also distilling an ensemble into a verifier. We instead train an answer generator, and our gain is mainly attributable to direction rather than to a new aggregation estimator.

\paragraph{Distilling selected answers.}
Sequence, on-policy, and best-of-$N$ distillation construct targets from teacher or verifier information \citep{kim2016sequence,agarwal2023gkd,sessa2024bond}. Reward-filtered self-training similarly selects training signals before updating a model \citep{dong2023raft,gulcehre2023rest,singh2023restem,answerconditioned2026}. Multi-teacher distillation and cross-tokenizer likelihood scoring provide further ways to extract supervision from model evidence \citep{jin2026exploring,phan2025crosstokenizer}. Our factual setting separates candidate coverage, label correctness, and student reproduction, which shows whether a gain comes from better targets or from better expression of the same targets.

\section{Conclusion and Limitations}
\label{sec:limits}
\label{app:limitations}
We introduced directional label distillation, which turns a teacher's ability to verify a relation in one direction into training targets for the other. Known-direction labels are more accurate than reverse labels under every aggregation rule and stronger comparator we tested, on screened and unscreened queries and without inserted answers, and students reproduce these labels almost exactly. Verifying candidates in the direction a model knows is therefore a practical source of supervision when generation fails. On our facts the better direction coincides with the notable entity rather than with the prompt template, so the entity that text describes more often is a natural criterion for choosing the scoring direction.

We study one relation over a fixed name inventory, which lets us control candidate sets, forward exposure, and evaluation exactly; relations with open-ended answers will need other ways to propose candidates. Scoring also costs one teacher call per candidate, so a larger inventory needs a cheaper proposal stage in front of the scorer. Evaluating students on the queries whose labels they learn isolates the contribution of label quality to the answers they give. Our corpus fixes the scoring direction by construction, and a label-free rule for choosing it is the natural next step for fact sets whose notable entity varies.

\section*{Ethics Statement}
All authors have read and adhere to the ICLR Code of Ethics. The study uses public Wikidata parent--child relations and publicly released language models. It involves no human subjects, annotators, or personal data beyond these public records. Scoring a relation in the direction a model can verify could also surface memorized relations about real people; we therefore restrict every experiment to documented Wikidata facts and evaluate against them without collecting new information about any individual. The notable-parent cohort includes people without English Wikipedia articles; for them we use only relations already recorded in Wikidata and report aggregate accuracies.

\bibliographystyle{iclr2027_conference}
\bibliography{refs}
\clearpage

\appendix
\renewcommand{\topfraction}{0.9}\renewcommand{\textfraction}{0.08}\renewcommand{\floatpagefraction}{0.85}
\setcounter{topnumber}{3}\setcounter{totalnumber}{4}
\section{Implementation and supervision}
\label{app:implementation}
\paragraph{Corpus.}
We mined the parent--child pool from the public Wikidata Query Service on July 12, 2026. Each query selects humans (P31 = Q5) with an English Wikipedia article and one of $19$ occupations (P106), together with their human fathers or mothers (P22, P25), and returns English labels. Names are accent-stripped and kept when they consist of two to four ASCII tokens. Pairs are deduplicated, and a parent query accepts every child recorded with that parent. Entities are identified by their cleaned names, and the stored pool of $10{,}505$ pairs defines the corpus.

\paragraph{Prompts and checkpoints.}
The known-direction prefixes are \texttt{\{child\}'s parent is} and \texttt{\{child\} is the child of}. Both score the parent continuation. Reverse scoring uses \texttt{\{parent\}'s child is} and scores the candidate child. Scores use mean continuation-token log probabilities with a leading space. Candidate sources are \texttt{Qwen/Qwen3-8B}, \texttt{allenai/OLMo-2-1124-7B}, and \texttt{mistralai/Mistral-7B-v0.3}. \texttt{meta-llama/Llama-3.1-8B-Instruct} rescores their pool. The ten- and twelve-channel variants add two templates each from \texttt{01-ai/Yi-1.5-9B} and \texttt{zai-org/GLM-Z1-9B-0414}. Matched direction comparisons use both directions from the same rescoring run.

\paragraph{Soft agreement and score means.}
Channel $r$ denotes a teacher--template pair and produces $P_{ir}(c)=\operatorname{softmax}_{c\in\mathcal C_i}\kap_r(p_i,c)$ at temperature one, after masking self-candidate scores to $-10^9$. With $M_i=|\mathcal C_i|$ counting stored slots, soft agreement combines channels into the candidate distribution $q_i(c)$ using weights $a_r$ and iterates

\begin{align}
q_i(c)&=\frac{1}{Z_i}\prod_r\left[\frac{1-a_r}{M_i}+a_rP_{ir}(c)\right],
\label{eq:posterior}\\
a_r&\leftarrow\operatorname{clip}_{[0.001,0.999]}
\left(\frac{1}{n}\sum_i\sum_{c\in\mathcal C_i}q_i(c)P_{ir}(c)\right).
\label{eq:agreement}
\end{align}

$Z_i$ normalizes $q_i$ over the stored candidates $\mathcal C_i$, and $n=1{,}500$ is the number of query distributions used to fit the weights. The product spans all channels. Weights start at $0.7$ and stop after $200$ updates or a maximum change below $10^{-9}$. The highest-posterior eligible candidate becomes the label. The masked self slot remains in the mixture floor. This overlap update is a heuristic inspired by latent-annotator aggregation \citep{dawid1979maximum,ratner2016data} rather than a Dawid--Skene likelihood estimate.

Raw-score means average scores directly. z-scores use each channel's mean and standard deviation over the stored candidate array, adding $10^{-9}$ to the denominator. Raw and z-score rules apply the self mask after aggregation. Probability means mask before channel softmax. Borda sums $M_{\mathrm{eligible}}-\mathrm{rank}$, where $M_{\mathrm{eligible}}$ counts eligible candidates, and RRF sums $1/(60+\mathrm{rank})$, with one-based ranks. Stable sorting and argmax ties preserve candidate order. All final selections exclude self candidates, and $86$ of the stored pools contain a self slot.

\paragraph{Masked-diffusion training.}
Qwen3-4B or Qwen3-0.6B supplies the initial weights. The mask-token embedding is initialized from the mean embedding, and an additive attention mask allows bidirectional attention over non-pad keys. Training uses bf16 and LoRA rank $64$, scale $128$, dropout $0.05$, on q/k/v/o/gate/up/down projections. Answers occupy ten truncated or padded slots. Each example concatenates the prompt, the answer name over those slots, and a final period; prompt and period tokens stay visible and unscored, while pads can be supervised. For example $i$, a uniformly sampled subset $S_i$ of $k_i$ slots, $k_i\sim\mathcal U\{1,\ldots,10\}$, is masked. 

Let $a_{ij}$ be the target token in answer slot $j$, $\tilde x_i$ the full input with slots $S_i$ replaced by mask tokens, and $\theta$ the student parameters. For minibatch $\mathcal B$, the masked-answer loss is

\begin{equation}
\mathcal L_{\mathrm{MDM}}=-\frac{\sum_{i\in\mathcal B}\sum_{j\in S_i}\log P_\theta(a_{ij}\mid\tilde x_i)}{\sum_{i\in\mathcal B}|S_i|}.
\label{eq:mdm_loss}
\end{equation}

AdamW uses learning rate $10^{-4}$, batch size $16$, gradient clipping $1.0$, and no schedule or warmup. Warm and mixed-SFT budgets are $4{,}000$ and $4{,}800$ steps. Writing the forward and reverse corpora as $D_{\mathrm{fwd}}$ and $D_{\mathrm{rev}}$, the mixed corpus repeats reverse examples $\max(1,\operatorname{round}(|D_{\mathrm{fwd}}|/|D_{\mathrm{rev}}|))$ times and concatenates forward examples, with seven copies under default exposure, five under the withheld-child recipe, and nineteen for students on the unscreened cohort. Greedy confidence-first inference fills ten slots in ten passes with dropout disabled.

\paragraph{AR budget.}
The Qwen3-8B AR control retains causal attention, uses answer-span causal loss, the same LoRA settings and optimizer, gradient checkpointing, and no gradient clipping. Nominal steps divide the MDM ten-slot token budget by measured AR answer length (about $5.4$ tokens). MDM examples supervise about $5.5$ slots on average, so AR receives approximately $1.8$ times as many realized supervised tokens. The MDM-4B/AR-8B comparison changes size, objective, and budget. Mean answer lengths $\bar L_{\mathrm{warm}}$ and $\bar L_{\mathrm{SFT}}$, including the period, are estimated from each phase's first $400$ examples. The AR step budgets are $\operatorname{round}(640{,}000/(16\bar L_{\mathrm{warm}}))$ and $\operatorname{round}(768{,}000/(16\bar L_{\mathrm{SFT}}))$. Eight-channel AR warm steps are $7{,}316$, $7{,}431$, and $7{,}445$ for runs A, B, and C, respectively, with $8{,}751$ SFT steps each.

\paragraph{Name matching.}
Open accuracy uses accent- and case-normalized full-name substring matching against any true child, without bare-surname credit. The postprocessor first gathers inventory entries sharing a lowercase whitespace-delimited token with the output, or uses the full inventory if none share a token. It selects the entry with the highest character-level sequence-matching ratio. The matched students use the sorted $8{,}218$-name inventory, and a fixed inventory order breaks ties. Empty outputs remain empty. Because this postprocessing follows generation, it can exploit name cues, which the inventory-enlargement control tests.

\paragraph{Screen and development access.}
The Qwen3-8B main-template screen scores every true parent against fifteen non-parent distractors, including up to three with the child's surname. It retains the highest-scoring true parent only when it strictly beats every distractor. Child order uses a fixed pseudorandom shuffle, and the screen covered the $4{,}400$ children at positions $2{,}400$--$6{,}799$ of that order, of which the $2{,}476$ passing facts, $56.3\%$ of those screened, form the cohort. Gold facts supply the screen, forward adaptation, configuration evaluation, and answer insertion into the unscreened candidate lists. For non-oracle student arms, label selection does not use information about which candidate is correct, and reverse SFT uses the selected pseudo-label. Soft weights use all $1{,}500$ unlabeled score distributions.

\section{Matched label comparisons}
\label{app:matched}

Tables~\ref{tab:directionfull}--\ref{tab:fusionfull} compare scoring directions, context correction, and known-direction aggregation using any-child correctness on the fixed acquired pool. Both directions are scored in the same rescoring run over all $28{,}463$ stored candidate slots.

\begin{table}[htbp]
\caption{Direction control on the $1{,}500$ acquired pools (\%, four models, main template). Known scores use mean tokens, and DC subtracts the summed context-only score. The last columns compare known with summed DC in points, with paired parent-cluster $95\%$ intervals, wins/losses, and Holm-corrected McNemar $p$.}
\label{tab:directionfull}
\centering\small
\setlength{\tabcolsep}{3pt}
\begin{tabular*}{\linewidth}{@{\extracolsep{\fill}}lrrrrlrr@{}}
\toprule
 & & \multicolumn{3}{c}{\tblhead{Reverse}} & \multicolumn{3}{c}{\tblhead{Known $-$ DC, summed}}\\
\cmidrule(lr){3-5}\cmidrule(l){6-8}
\tblhead{Aggregation} & \tblhead{Known} & \tblhead{Mean} & \tblhead{Summed} & \tblhead{DC, sum} & \tblhead{Points [95\% CI]} & \tblhead{W/L} & \tblhead{Holm $p$}\\
\midrule
\textbf{Raw-score mean} & \cellcolor{oursrow}90.73 & 54.60 & 72.40 & 75.60 & 15.13 $[12.96,17.32]$ & 259/32 & $1.0\times10^{-44}$\\
\tblindent{z-score mean} & 90.27 & 54.27 & 72.40 & 76.20 & 14.07 $[11.93,16.28]$ & 248/37 & $4.8\times10^{-39}$\\
\tblindent{Probability mean} & 90.87 & 54.27 & 73.00 & 74.13 & 16.73 $[14.54,18.96]$ & 280/29 & $6.0\times10^{-52}$\\
\tblindent{Borda} & 88.93 & 55.20 & 72.07 & 76.47 & 12.47 $[10.38,14.60]$ & 226/39 & $3.4\times10^{-33}$\\
\tblindent{RRF} & 89.13 & 54.53 & 71.60 & 76.00 & 13.13 $[11.02,15.30]$ & 236/39 & $3.1\times10^{-35}$\\
\tblindent{Soft agreement} & 90.87 & 54.33 & 73.27 & 75.00 & 15.87 $[13.72,18.07]$ & 266/28 & $4.0\times10^{-49}$\\
\bottomrule
\end{tabular*}
\end{table}

\begin{table}[htbp]
\caption{Domain-context correction on all trained queries and their within-pair surname-mismatched subset. Accuracy is percent.}
\label{tab:dc}
\centering\small
\begin{tabular}{lrrrrrr}
\toprule
 & \multicolumn{3}{c}{\tblhead{All $1{,}500$}} & \multicolumn{3}{c}{\tblhead{Surname mismatch, $305$}}\\
\cmidrule(lr){2-4}\cmidrule(lr){5-7}
\tblhead{Mean} & \tblhead{Known} & \tblhead{Reverse} & \tblhead{DC} & \tblhead{Known} & \tblhead{Reverse} & \tblhead{DC}\\
\midrule
\textbf{Raw} & \cellcolor{oursrow}90.73 & 54.60 & 55.60 & \cellcolor{oursrow}71.80 & 40.66 & 25.25\\
\tblindent{z-score} & 90.27 & 54.27 & 56.40 & 70.82 & 40.33 & 25.90\\
\tblindent{Probability} & 90.87 & 54.27 & 53.87 & 72.79 & 40.98 & 23.61\\
\bottomrule
\end{tabular}
\end{table}

Table~\ref{tab:dc} applies context correction with \texttt{The child is} and coefficient one to mean-token scores. All four background maps cover the $7{,}062$ distinct candidate names. Each conditional term is normalized separately by its continuation-token count, giving a PMI-like contrast rather than calibrated joint-distribution PMI. The surname stratum compares lowercase final whitespace tokens within each recorded parent--child pair.

Tables~\ref{tab:directionfull} and~\ref{tab:dc} compare label accuracy with summed-token and mean-token DC reverse controls, respectively, on all $1{,}500$ trained queries. The results for raw-score means show that known-direction labels are more accurate than mean-token DC reverse labels. The accuracy difference has a parent-cluster interval of $[32.47,37.78]$ points, with $557$ wins and $30$ losses. For mean-token DC versus uncorrected reverse labels, the interval is $[-2.58,4.63]$ points and includes zero. On the $1{,}390$ exposure-free queries, known minus summed DC is $12.95$ to $16.83$ points across the six rules; the rules span $1.87$ points for the known direction and $2.01$ points for summed DC. On all $1{,}500$ trained queries the same spans are $1.93$ and $2.33$ points.

These comparisons use exact McNemar tests and paired correctness differences. Mean-token contrasts use $20{,}000$ row or parent-cluster bootstrap replicates and percentile intervals, where cluster resampling draws $1{,}467$ parents and divides summed differences by summed row counts. Summed-token contrasts use $10{,}000$ parent-cluster draws.

\begin{table}[htbp]
\caption{Known-direction aggregation on the acquired pool. The first six rules combine the same eight score vectors, and the last two rows average only Llama's two template vectors.}
\label{tab:fusionfull}
\centering\small
\begin{tabular}{lrr}
\toprule
\tblhead{Rule} & \tblhead{Vectors} & \tblhead{Accuracy (\%)}\\
\midrule
\rowcolor{oursrow}[0pt][0pt]
\textbf{Soft agreement} & 8 & 91.53\\
Raw-score mean & 8 & 91.20\\
z-score mean & 8 & 91.27\\
Probability mean & 8 & 91.40\\
Borda & 8 & 89.93\\
Reciprocal-rank fusion & 8 & 90.07\\
\midrule
Llama two-template raw mean & 2 & 90.53\\
Llama two-template probability mean & 2 & 90.60\\
\bottomrule
\end{tabular}
\end{table}

Table~\ref{tab:fusionfull} compares aggregation rules over known-direction scores. Six prespecified comparisons apply Holm correction to agreement versus the three score means, Borda, RRF, and Llama raw mean. Agreement and the three score means differ by less than one point.

\paragraph{Name-prior and length strata.}
\label{app:strata}
Table~\ref{tab:strata} lists the strata behind Figure~\ref{fig:nameprior}, and Table~\ref{tab:strataerrors} counts the wrong selections that prefer a more probable name. Known-direction accuracy is about $94\%$ on the acquired-pool subset of $1{,}346$ exposure-free queries whose recorded child is among the stored candidates. The other $44$ of the $1{,}390$ exposure-free queries lack the recorded child in their stored candidates. The acquired-pool analysis therefore measures label selection when the recorded answer is available.

A candidate's name score is its summed context-only log probability after \texttt{The child is}, averaged over the four teachers, and its continuation length is the mean token count of the child name across their tokenizers. Terciles are computed within each pool; the two unscreened pools share true children and therefore cut points. Intervals use $10{,}000$ paired bootstrap resamples of parents.

\begin{table}[htbp]
\caption{Label accuracy (\%) by tercile of the true child's context-only name score and continuation length. K is the mean-token known direction, R and D sum reverse and DC tokens, and K $-$ D carries a $95\%$ interval. Blue marks known-direction labels.}
\label{tab:strata}
\centering\small
\begin{tabular*}{\linewidth}{@{\extracolsep{\fill}}llrrrrl@{}}
\toprule
\tblhead{Pool} & \tblhead{Tercile} & \tblhead{$n$} & \tblhead{K} & \tblhead{R} & \tblhead{D} & \tblhead{K $-$ D [95\% CI]}\\
\midrule
\rowcolor{groupbg}[0pt][0pt]\multicolumn{7}{@{}c@{}}{\textbf{Context-only name score}\strut}\\
Acquired & Low & 449 & \cellcolor{oursrow}95.1 & 61.7 & 89.8 & 5.3 $[2.7,8.1]$\\
 & Mid & 448 & \cellcolor{oursrow}92.2 & 78.1 & 85.7 & 6.5 $[3.1,10.0]$\\
 & High & 449 & \cellcolor{oursrow}93.5 & 82.4 & 57.5 & 36.1 $[31.4,40.7]$\\
\addlinespace[1.5pt]
Uniform & Low & 171 & \cellcolor{oursrow}90.6 & 49.7 & 86.5 & 4.1 $[0.6,8.2]$\\
 & Mid & 170 & \cellcolor{oursrow}82.4 & 61.2 & 70.0 & 12.4 $[6.5,18.8]$\\
 & High & 171 & \cellcolor{oursrow}90.6 & 69.6 & 63.7 & 26.9 $[19.9,33.9]$\\
\addlinespace[1.5pt]
Lexical & Low & 171 & \cellcolor{oursrow}75.4 & 40.4 & 74.9 & 0.6 $[-4.7,5.8]$\\
 & Mid & 170 & \cellcolor{oursrow}63.5 & 46.5 & 53.5 & 10.0 $[3.5,16.5]$\\
 & High & 171 & \cellcolor{oursrow}74.9 & 53.8 & 39.2 & 35.7 $[28.1,43.3]$\\
\midrule
\rowcolor{groupbg}[0pt][0pt]\multicolumn{7}{@{}c@{}}{\textbf{Continuation length}\strut}\\
Acquired & Low & 568 & \cellcolor{oursrow}93.0 & 81.0 & 76.2 & 16.7 $[13.1,20.4]$\\
 & Mid & 363 & \cellcolor{oursrow}94.2 & 76.0 & 77.7 & 16.5 $[12.1,21.0]$\\
 & High & 415 & \cellcolor{oursrow}94.0 & 62.9 & 79.5 & 14.5 $[10.8,18.3]$\\
\addlinespace[1.5pt]
Uniform & Low & 201 & \cellcolor{oursrow}84.1 & 66.7 & 64.7 & 19.4 $[12.9,25.9]$\\
 & Mid & 143 & \cellcolor{oursrow}86.7 & 61.5 & 75.5 & 11.2 $[6.3,16.8]$\\
 & High & 168 & \cellcolor{oursrow}93.5 & 51.2 & 82.1 & 11.3 $[6.5,16.7]$\\
\addlinespace[1.5pt]
Lexical & Low & 201 & \cellcolor{oursrow}65.2 & 49.3 & 43.8 & 21.4 $[14.4,28.4]$\\
 & Mid & 143 & \cellcolor{oursrow}70.6 & 48.3 & 58.7 & 11.9 $[4.9,18.9]$\\
 & High & 168 & \cellcolor{oursrow}79.2 & 42.9 & 67.9 & 11.3 $[4.8,17.9]$\\
\bottomrule
\end{tabular*}
\end{table}

\begin{table}[htbp]
\caption{Wrong label selections whose context-only name score exceeds the true child's, counted over all wrong selections of each score. Blue marks known-direction labels.}
\label{tab:strataerrors}
\centering\small
\begin{tabular}{lrrrr}
\toprule
\tblhead{Pool} & \tblhead{$n$} & \tblhead{K} & \tblhead{R} & \tblhead{D}\\
\midrule
Acquired & 1,346 & \cellcolor{oursrow}40/86 & 315/349 & 38/301\\
Uniform & 512 & \cellcolor{oursrow}15/62 & 199/204 & 19/136\\
Lexical & 512 & \cellcolor{oursrow}57/147 & 245/272 & 68/226\\
\bottomrule
\end{tabular}
\end{table}

\section{Matched forward-withheld students}
\label{app:newstudents}
\paragraph{Inputs and hardware.}
Five label sources use the same ordered $1{,}500$ training records. Known, reverse, and DC reverse use the four-teacher main-template scores of the rescoring run. Llama and agreement labels combine main-template scores from an earlier pass over the same pool, which differ from the rescoring run by $0.005$ to $0.022$ in mean absolute mean-token log probability, with alternate-template scores from that run. Removing $2{,}617$ forward edges involving the children of the first $2{,}000$ shuffled facts leaves $7{,}888$ edges for both stages. The primary $1{,}390$ trained pairs have no query parent in that retained corpus, excluding direct and sibling answer exposure. All $1{,}500$ reverse examples remain in training.

Within the primary student comparison, run A uses A100 GPUs and runs B and C use RTX A6000 GPUs. GPU type is matched across label sources within each run.

\paragraph{Checkpoint and randomness.}
Qwen3-0.6B uses a BF16 base, FP32 LoRA parameters, and a fresh optimizer per phase. Adapters omit embedding and output-head weights, and reloading recreates the mean-initialized mask-token rows. The run seed controls parameter initialization, dropout, forward shuffling, probe sampling, and evaluation order, whereas the batch-index and masking generators reset to the same fixed initial state at the start of each phase. All runs use PyTorch 2.11.0, Transformers 5.12.1, PEFT 0.19.1, and Accelerate 1.14.0.

\paragraph{Evaluation and whole-answer equality.}
\label{app:exactanswer}
Table~\ref{tab:newprimary} reports student accuracy and exact counts of whole-answer correctness and label fidelity for each training seed. Open and whole-answer accuracy use the same generated answers. For whole-answer evaluation, we apply Unicode NFKD decomposition, remove non-ASCII characters and punctuation, lowercase, and collapse whitespace. A correct answer must exactly match a nonempty true-child name after normalization. Label fidelity requires the same exact match to the selected pseudo-label, whether that label is correct or incorrect. We summarize variation across these three seeds using sample standard deviations.

\begin{table}[htbp]
\caption{Primary student results by training run. Open and matched entries are percentages. Whole-answer and label-equality entries are exact counts out of $1{,}390$.}
\label{tab:newprimary}
\centering\small
\begin{tabular}{llrrrr}
\toprule
\tblhead{Labels} & \tblhead{Run} & \tblhead{Open (\%)} & \tblhead{Whole count} & \tblhead{Matched (\%)} & \tblhead{Label count}\\
\midrule
\multirow{3}{*}{\textbf{Known mean}} & A & \cellcolor{oursrow}80.29 & \cellcolor{oursrow}1112 & \cellcolor{oursrow}89.28 & \cellcolor{oursrow}1211\\
 & B & \cellcolor{oursrow}78.99 & \cellcolor{oursrow}1095 & \cellcolor{oursrow}89.14 & \cellcolor{oursrow}1186\\
 & C & \cellcolor{oursrow}74.82 & \cellcolor{oursrow}1035 & \cellcolor{oursrow}88.42 & \cellcolor{oursrow}1129\\
\midrule
\multirow{3}{*}{Reverse mean} & A & 48.63 & 673 & 53.60 & 1132\\
 & B & 46.69 & 648 & 53.02 & 1103\\
 & C & 46.19 & 639 & 52.81 & 1084\\
\midrule
\multirow{3}{*}{DC reverse mean} & A & 49.21 & 683 & 53.53 & 1256\\
 & B & 49.35 & 682 & 53.81 & 1252\\
 & C & 49.21 & 681 & 54.46 & 1260\\
\midrule
\multirow{3}{*}{Llama two-template} & A & 78.71 & 1090 & 88.92 & 1192\\
 & B & 74.89 & 1037 & 88.35 & 1128\\
 & C & 71.73 & 995 & 87.41 & 1083\\
\midrule
\multirow{3}{*}{Agreement (8)} & A & 78.35 & 1086 & 89.86 & 1170\\
 & B & 78.99 & 1097 & 89.78 & 1178\\
 & C & 79.93 & 1102 & 89.71 & 1195\\
\bottomrule
\end{tabular}
\end{table}

\paragraph{Surname mismatch.}
\label{app:scope}
The $294$-pair trained intersection requires different parent and child last-name tokens within each recorded pair, in addition to the primary exposure exclusion, and Figure~\ref{fig:namecues}a reports the known-direction and summed-DC students on it.

\paragraph{Summed-token domain-context control.}
Table~\ref{tab:summeddcstudents} compares the open, whole-answer, and inventory-matched accuracy of students trained on known-direction and summed-token DC labels. The summed-token DC control uses the existing reverse and background scores, with training and evaluation matched to the primary comparison. In this comparison, known-direction labels are correct for $1{,}261/1{,}390$ primary queries ($90.72\%$), compared with $1{,}045/1{,}390$ ($75.18\%$) for summed DC, a $15.54$-point gap. The results show that this label advantage persists after training, with known-direction students achieving higher open and whole-answer accuracy in all three seeds.

\begin{table}[!htb]
\caption{Summed-token DC reverse control on the $1{,}390$ exposure-free trained queries. Accuracies are percentages and paired gains are percentage points; summary rows give three-seed means $\pm$ sample SD.}
\label{tab:summeddcstudents}
\centering\small
\begin{tabular}{llrrr}
\toprule
\tblhead{Label source} & \tblhead{Run} & \tblhead{Open} & \tblhead{Whole answer} & \tblhead{Matched}\\
\midrule
Known-direction mean & A & 80.29 & 80.00 & 89.28\\
 & B & 78.99 & 78.78 & 89.14\\
 & C & 74.82 & 74.46 & 88.42\\
\rowcolor{oursrow}[0pt][0pt]
 & Mean & $78.03\pm2.86$ & $77.75\pm2.91$ & $88.94\pm0.46$\\
\midrule
DC reverse sum & A & 63.17 & 63.09 & 73.81\\
 & B & 64.60 & 64.53 & 73.88\\
 & C & 62.01 & 61.80 & 74.24\\
 & Mean & $63.26\pm1.30$ & $63.14\pm1.37$ & $73.98\pm0.23$\\
\midrule
Known minus DC & Paired & $14.77\pm2.18$ & $14.60\pm2.15$ & $14.96\pm0.69$\\
\bottomrule
\end{tabular}
\end{table}

\section{Selected-label transfer}
\label{app:labelinternalization}
Table~\ref{tab:labelinternalization} reports selected-label accuracy and student reproduction for the primary directions, the summed-token DC comparator, and the exposure-free disagreement subset of the primary directions. Label reproduction and error correction were preregistered before student training, and the exposure-free and label-disagreement subsets are descriptive stratifications. Inventory identity $I$ requires exact equality to the selected pseudo-label. Open inclusion $O$ credits its nonempty normalized full name within the generated string.

\begin{table}[htbp]
\caption{Selected-label accuracy, inventory identity, and open inclusion, in percent. Identity and inclusion report mean $\pm$ sample SD. Label accuracy is fixed.}
\label{tab:labelinternalization}
\centering\small
\begin{tabular}{lrrr}
\toprule
\tblhead{Labels} & \tblhead{Label accuracy} & \tblhead{$I$} & \tblhead{$O$}\\
\midrule
\rowcolor{groupbg}[0pt][0pt]\multicolumn{4}{@{}c@{}}{\textbf{Exposure-free queries, $n=1{,}390$}\strut}\\
\rowcolor{oursrow}[0pt][0pt]
\textbf{Known mean} & $90.72$ & $97.70\!\pm\!0.61$ & $84.82\!\pm\!2.97$\\
Reverse mean & $53.60$ & $97.67\!\pm\!0.30$ & $80.12\!\pm\!1.63$\\
DC reverse sum & $75.18$ & $97.39\!\pm\!0.51$ & $80.24\!\pm\!1.30$\\
\midrule
\rowcolor{groupbg}[0pt][0pt]\multicolumn{4}{@{}c@{}}{\textbf{Exposure-free label-disagreement queries, $n=666$}\strut}\\
\rowcolor{oursrow}[0pt][0pt]
\textbf{Known mean} & $82.73$ & $97.25\!\pm\!0.76$ & $84.33\!\pm\!2.23$\\
Reverse mean & $5.26$ & $96.60\!\pm\!0.35$ & $72.12\!\pm\!1.00$\\
\bottomrule
\end{tabular}
\end{table}

Table~\ref{tab:labelinternalizationseeds} reports selected-label reproduction and changes in correctness under matched and open scoring. An additional whole-answer comparison finds that the known-direction, reverse, and summed-DC students never turn a wrong label into a whole-answer match in any seed. Label accuracy is therefore an upper bound on whole-answer accuracy in these runs.

For the matched and open counts in the table, $F$ counts wrong labels with correct student answers and $L$ counts correct labels with wrong answers. Subscripts distinguish matched and open scoring. For $N_+$ correct labels, the four correctness cells are $N_+-L$, $F$, $L$, and $n-N_+-F$. Student minus label accuracy is $(F-L)/n$. Primary $N_+$ is $1261$ for known-direction labels, $745$ for reverse labels, and $1045$ for summed-DC labels.

\begin{table}[htbp]
\caption{Primary selected-label transfer counts by training run. $I/O$ count identity/inclusion. $F/L$ count answers fixed or lost against the selected label under matched ($m$) and open ($o$) scoring.}
\label{tab:labelinternalizationseeds}
\centering\small
\begin{tabular}{llrrrrrr}
\toprule
\tblhead{Labels} & \tblhead{Run} & \tblhead{$I$} & \tblhead{$O$} & \tblhead{$F_m$} & \tblhead{$L_m$} & \tblhead{$F_o$} & \tblhead{$L_o$}\\
\midrule
\multirow{3}{*}{\textbf{Known mean}} & A & \cellcolor{oursrow}1366 & \cellcolor{oursrow}1215 & \cellcolor{oursrow}0 & \cellcolor{oursrow}20 & \cellcolor{oursrow}1 & \cellcolor{oursrow}146\\
 & B & \cellcolor{oursrow}1359 & \cellcolor{oursrow}1188 & \cellcolor{oursrow}1 & \cellcolor{oursrow}23 & \cellcolor{oursrow}1 & \cellcolor{oursrow}164\\
 & C & \cellcolor{oursrow}1349 & \cellcolor{oursrow}1134 & \cellcolor{oursrow}1 & \cellcolor{oursrow}33 & \cellcolor{oursrow}0 & \cellcolor{oursrow}221\\
\multirow{3}{*}{Reverse mean} & A & 1361 & 1138 & 5 & 5 & 0 & 69\\
 & B & 1359 & 1110 & 4 & 12 & 0 & 96\\
 & C & 1353 & 1093 & 3 & 14 & 0 & 103\\
\multirow{3}{*}{DC reverse sum} & A & 1353 & 1098 & 3 & 22 & 1 & 168\\
 & B & 1347 & 1134 & 5 & 23 & 1 & 148\\
 & C & 1361 & 1114 & 4 & 17 & 0 & 183\\
\bottomrule
\end{tabular}
\end{table}

\section{Cohort with direction-independent candidates}
\label{app:prospective}
\paragraph{Cohort and pools.}
We fixed the cohort and both policies before inference, without requiring the forward screen. Excluding every parent query and recorded child of those first $2{,}000$ screened facts leaves $7{,}413$ eligible parents. A pseudorandom order, fixed by a salted hash, selects $512$ distinct parents and distinct recorded children without inspecting teacher scores or screen outcomes. Each pool has $64$ names and excludes the query entity. Uniform pools contain the recorded child and $63$ pseudorandom inventory names. Lexical pools contain the recorded child, the $32$ names with greatest character-trigram Jaccard similarity to the parent, and $31$ further pseudorandom names. Similarity uses accent- and case-normalized strings. Hashes fix ties and positions. Answer insertion ensures coverage. Other sampled valid children also receive credit, and the policies reuse the same $512$ queries.

\paragraph{Scoring.}
The four frozen teachers score raw prompts without a chat wrapper. Each scores all $65{,}536$ query--candidate slots in both directions using the main templates and an exact prompt-token-prefix check. Context scores after \texttt{The child is} are computed once per distinct candidate. Four-model means fit no weights. Correctness requires exact candidate-name membership in the reference child set. All slots were scorable, with no post-inference removal. Summed-token scoring separately replaces mean continuations with unnormalized sums and uses the corresponding summed context subtraction.

The primary tests compare the four-model known-direction mean with raw and corrected reverse means within each pool. Intervals use $10{,}000$ paired bootstrap resamples of parent queries. Exact McNemar tests receive Holm correction across four prespecified contrasts. Per-teacher, summed-score, and surname analyses are secondary.

\paragraph{Alternate template pair.}
Table~\ref{tab:prospectiveteachers} compares label accuracy across teachers and both template pairs, and Table~\ref{tab:prospectivepaired} reports paired known-direction gains over reverse scores. The preregistered alternate pair scores the parent after \texttt{\{child\} is the child of} and the candidate child after \texttt{\{parent\} is the parent of}, with \texttt{Someone is the parent of} as the context-only background. We use the same cohort, both candidate pools, and all four teachers, with an exact prompt-token prefix check for all $65{,}536$ query--candidate pairs scored by each teacher. Using four-teacher means, we compare mean-token known-direction labels against summed reverse and summed DC labels in each pool. These four preregistered comparisons use exact McNemar tests with Holm correction; confidence intervals use $10{,}000$ paired bootstrap resamples of parent queries. The results show that known-direction labels are more accurate than summed-DC labels in both pools. Averaging both templates for each teacher into eight channels gives gains over summed DC of $13.28$ $[9.96,16.80]$ and $15.62$ $[11.72,19.53]$ points. Within one teacher, mean-token and summed-token known-direction rankings coincide because the scored parent continuation is identical for every candidate. All preregistered contrasts have Holm-corrected $p<10^{-4}$, and the main-template summed-token contrasts are descriptive. Table~\ref{tab:prospectivesurname} compares scoring directions on the $228$ within-pair surname-mismatched queries. In the uniform pool, the known direction reaches $73.25\%$ under both templates, against at most $44.30\%$ for summed DC.

\begin{table}[htbp]
\caption{Per-teacher candidate accuracy (\%) on the unscreened cohort under both template pairs. C$\to$P is the known direction, scoring the query parent after each candidate child, and P$\to$C scores each candidate child after the parent.}
\label{tab:prospectiveteachers}
\centering\small
\begin{tabular*}{\linewidth}{@{\extracolsep{\fill}}llrrrrrr@{}}
\toprule
 & & \multicolumn{3}{c}{\tblhead{Mean tokens}} & \multicolumn{3}{c}{\tblhead{Summed tokens}}\\
\cmidrule(lr){3-5}\cmidrule(l){6-8}
\tblhead{Template} & \tblhead{Scorer} & \tblhead{C$\to$P} & \tblhead{P$\to$C} & \tblhead{DC} & \tblhead{C$\to$P} & \tblhead{P$\to$C} & \tblhead{DC}\\
\midrule
\rowcolor{groupbg}[0pt][0pt]\multicolumn{8}{@{}c@{}}{\textbf{Uniform pool, $512$ queries}\strut}\\
Main & Qwen3-8B & 72.27 & 23.44 & 47.27 & 72.27 & 52.15 & 57.03\\
 & OLMo-2-7B & 81.05 & 32.81 & 50.20 & 81.05 & 55.08 & 64.06\\
 & Mistral-7B & 84.38 & 31.84 & 61.52 & 84.38 & 61.72 & 70.31\\
 & Llama-3.1-8B & 88.48 & 26.37 & 41.60 & 88.48 & 55.66 & 62.50\\
 & \textbf{Four-model mean} & \cellcolor{oursrow}87.89 & 32.23 & 61.33 & \cellcolor{oursrow}87.50 & 60.16 & 73.44\\
\addlinespace[2pt]
Alternate & Qwen3-8B & 73.05 & 35.55 & 36.91 & 73.05 & 54.30 & 48.05\\
 & OLMo-2-7B & 80.27 & 39.06 & 56.64 & 80.27 & 58.40 & 65.82\\
 & Mistral-7B & 84.77 & 35.16 & 63.67 & 84.77 & 60.16 & 65.82\\
 & Llama-3.1-8B & 87.89 & 36.91 & 62.70 & 87.89 & 60.55 & 67.19\\
 & \textbf{Four-model mean} & \cellcolor{oursrow}87.89 & 40.43 & 65.23 & \cellcolor{oursrow}87.89 & 60.55 & 72.46\\
\midrule
\rowcolor{groupbg}[0pt][0pt]\multicolumn{8}{@{}c@{}}{\textbf{Lexical pool, $512$ queries}\strut}\\
Main & Qwen3-8B & 56.45 & 24.41 & 31.45 & 56.45 & 39.26 & 43.55\\
 & OLMo-2-7B & 63.48 & 28.32 & 29.10 & 63.48 & 41.60 & 47.85\\
 & Mistral-7B & 65.62 & 30.08 & 41.60 & 65.62 & 45.51 & 55.66\\
 & Llama-3.1-8B & 75.59 & 26.17 & 27.54 & 75.59 & 45.51 & 51.76\\
 & \textbf{Four-model mean} & \cellcolor{oursrow}71.29 & 30.08 & 37.30 & \cellcolor{oursrow}71.29 & 46.88 & 55.86\\
\addlinespace[2pt]
Alternate & Qwen3-8B & 56.64 & 32.62 & 24.02 & 56.64 & 41.60 & 37.70\\
 & OLMo-2-7B & 62.50 & 33.98 & 36.33 & 62.50 & 42.97 & 48.63\\
 & Mistral-7B & 68.95 & 31.64 & 38.48 & 68.95 & 45.12 & 51.76\\
 & Llama-3.1-8B & 75.78 & 32.62 & 42.58 & 75.78 & 45.90 & 50.39\\
 & \textbf{Four-model mean} & \cellcolor{oursrow}72.07 & 35.35 & 43.16 & \cellcolor{oursrow}72.07 & 47.46 & 54.30\\
\bottomrule
\end{tabular*}
\end{table}

\begin{table}[htbp]
\caption{Paired known-direction gains on the unscreened cohort (points, four-model means) with $95\%$ bootstrap intervals and wins/losses. The known direction uses mean tokens throughout.}
\label{tab:prospectivepaired}
\centering\small
\begin{tabular*}{\linewidth}{@{\extracolsep{\fill}}lllrrr@{}}
\toprule
\tblhead{Template} & \tblhead{Comparator} & \tblhead{Pool} & \tblhead{Gain} & \tblhead{95\% interval} & \tblhead{Wins/losses}\\
\midrule
Main & P$\to$C, mean token & Uniform & 55.66 & $[51.37,59.96]$ & 285/0\\
 & DC, mean token & Uniform & 26.56 & $[22.46,30.86]$ & 145/9\\
 & P$\to$C, mean token & Lexical & 41.21 & $[36.91,45.51]$ & 214/3\\
 & DC, mean token & Lexical & 33.98 & $[29.10,38.67]$ & 193/19\\
\addlinespace[2pt]
Main & P$\to$C, summed token & Uniform & 27.73 & $[23.83,31.84]$ & 145/3\\
 & DC, summed token & Uniform & 14.45 & $[11.13,17.97]$ & 82/8\\
 & P$\to$C, summed token & Lexical & 24.41 & $[20.12,28.71]$ & 142/17\\
 & DC, summed token & Lexical & 15.43 & $[11.33,19.34]$ & 101/22\\
\addlinespace[2pt]
Alternate & P$\to$C, summed token & Uniform & 27.34 & $[23.44,31.45]$ & 143/3\\
 & DC, summed token & Uniform & 15.43 & $[12.30,18.75]$ & 81/2\\
 & P$\to$C, summed token & Lexical & 24.61 & $[20.31,28.91]$ & 140/14\\
 & DC, summed token & Lexical & 17.77 & $[13.87,21.68]$ & 105/14\\
\bottomrule
\end{tabular*}
\end{table}

\begin{table}[htbp]
\caption{Accuracy (\%) on $228$ within-pair surname-mismatched queries, using unchanged candidate lists. Per-teacher rows use the main template and mean tokens. The known direction (C$\to$P) uses mean tokens in every row; the last two rows sum reverse and DC tokens.}
\label{tab:prospectivesurname}
\centering\small
\begin{tabular}{lrrrrrr}
\toprule
 & \multicolumn{3}{c}{\tblhead{Uniform pool}} & \multicolumn{3}{c}{\tblhead{Lexical pool}}\\
\cmidrule(lr){2-4}\cmidrule(lr){5-7}
\tblhead{Scorer} & \tblhead{C$\to$P} & \tblhead{P$\to$C} & \tblhead{DC} & \tblhead{C$\to$P} & \tblhead{P$\to$C} & \tblhead{DC}\\
\midrule
\tblindent{Qwen3-8B} & 41.67 & 7.02 & 17.11 & 24.56 & 6.14 & 9.65\\
\tblindent{OLMo-2-7B} & 58.33 & 15.35 & 22.81 & 35.96 & 12.28 & 8.33\\
\tblindent{Mistral-7B} & 65.79 & 16.23 & 35.09 & 42.54 & 12.28 & 17.98\\
\tblindent{Llama-3.1-8B} & 75.44 & 15.79 & 23.25 & 58.33 & 14.47 & 12.72\\
\textbf{Four-model mean} & \cellcolor{oursrow}73.25 & 15.35 & 32.02 & \cellcolor{oursrow}47.81 & 12.28 & 13.60\\
\addlinespace[2pt]
Summed, main template & \cellcolor{oursrow}73.25 & 29.39 & 44.30 & \cellcolor{oursrow}47.81 & 18.42 & 25.44\\
Summed, alternate template & \cellcolor{oursrow}73.25 & 28.07 & 41.67 & \cellcolor{oursrow}48.68 & 17.11 & 23.68\\
\bottomrule
\end{tabular}
\end{table}

\paragraph{Full-inventory scans without inserted answers.} Table~\ref{tab:fullscan} compares candidate coverage and end-to-end label accuracy for preregistered full-inventory scans in the known and reverse directions, without inserted answers. Across teachers, known-direction scans place a true child among the top eight names for $63.67$ to $75.39\%$ of parents, against $26.37$ to $50.20\%$ for summed DC. The known-direction pipeline exceeds the summed-DC pipeline by $27.15$ $[22.85,31.45]$ points with $153$ wins and $14$ losses (Holm $p=2.0\times10^{-30}$), and on the union of both pools by $25.59$ $[21.29,29.88]$ points with $146$ wins and $15$ losses (Holm $p=3.8\times10^{-28}$). The shared-pool gain shows that the advantage extends beyond candidate retrieval to label selection.

For each of the $512$ query parents, the three acquisition teachers score all $8{,}218$ corpus children in both sentence directions using the scorer described above. Each scan covers $4{,}207{,}616$ query--name pairs, and each teacher also scores every inventory name after \texttt{The child is}. Total scoring cost grows linearly with inventory size and teacher count. Names equal to the query parent are ineligible, and pool construction uses no gold labels. Each pool unites the three teachers' top-eight names under mean-token known-direction, mean-token reverse, summed reverse, or summed DC scores. Llama-3.1-8B-Instruct scores every candidate in these pools without adding candidates, and each pipeline applies its four-teacher rule to its own pool.

We choose the reverse form with the highest end-to-end accuracy against gold labels on these queries as the primary comparator. A second contrast applies the known-direction and summed-DC rules to the union of their candidate pools. Intervals use $10{,}000$ paired bootstrap resamples of parents, and exact McNemar tests receive Holm correction across the two contrasts.

\begin{table}[htbp]
\caption{Full-inventory scans without inserted answers on the $512$ unscreened parents (coverage and accuracy in \%). (a) Pools unite three teachers' top-eight names under one score; coverage counts parents whose pool holds a true child. (b) Parents with a true child among one teacher's top $k$ names.}
\label{tab:fullscan}
\centering\small
\setlength{\tabcolsep}{3pt}
\textbf{(a) Pipelines}\par\smallskip
\begin{tabular*}{0.7\linewidth}{@{\extracolsep{\fill}}lrrr@{}}
\toprule
\tblhead{Pool and rule} & \tblhead{Mean size (names)} & \tblhead{Coverage} & \tblhead{Accuracy}\\
\midrule
\rowcolor{oursrow}[0pt][0pt]
\textbf{Known direction} & 19.33 & 80.66 & 66.21\\
Reverse, mean token & 19.53 & 21.09 & 5.27\\
Reverse, summed & 19.17 & 53.71 & 29.10\\
DC reverse, summed & 22.26 & 57.03 & 39.06\\
\addlinespace[1.5pt]
\rowcolor{oursrow}[0pt][0pt]
Union, known rule & 39.54 & 81.84 & 66.02\\
Union, summed-DC rule & 39.54 & 81.84 & 40.43\\
\bottomrule
\end{tabular*}
\par\bigskip
\textbf{(b) Per-teacher top-$k$ coverage}\par\smallskip
\begin{tabular*}{\linewidth}{@{\extracolsep{\fill}}lrrrrrrrrr@{}}
\toprule
 & \multicolumn{3}{c}{\tblhead{Qwen3-8B}} & \multicolumn{3}{c}{\tblhead{OLMo-2-7B}} & \multicolumn{3}{c}{\tblhead{Mistral-7B}}\\
\cmidrule(lr){2-4}\cmidrule(lr){5-7}\cmidrule(l){8-10}
\tblhead{Score} & \tblhead{$k{=}1$} & \tblhead{$8$} & \tblhead{$48$} & \tblhead{$1$} & \tblhead{$8$} & \tblhead{$48$} & \tblhead{$1$} & \tblhead{$8$} & \tblhead{$48$}\\
\midrule
\rowcolor{oursrow}[0pt][0pt]
\textbf{Known direction} & 49.02 & 63.67 & 69.92 & 55.66 & 72.46 & 78.52 & 55.86 & 75.39 & 81.45\\
Reverse, mean token & 2.93 & 9.96 & 19.53 & 6.84 & 14.06 & 27.93 & 5.66 & 12.89 & 25.59\\
Reverse, summed & 18.55 & 37.70 & 49.41 & 22.07 & 43.55 & 54.10 & 26.76 & 44.14 & 56.64\\
DC reverse, summed & 7.62 & 26.37 & 52.15 & 19.73 & 45.12 & 63.48 & 26.56 & 50.20 & 67.38\\
\bottomrule
\end{tabular*}
\end{table}

\paragraph{Students on the unscreened cohort.}
Table~\ref{tab:prospectivestudents} compares students trained on known-direction and summed-token DC labels on each unscreened pool. Across both pools, known-direction students have higher open, whole-answer, and inventory-matched accuracy in every seed, with paired gains of $12.30$ to $16.41$ points across seeds and the three metrics. No student turns a wrong label into a whole-answer match. In a descriptive stratification of the $228$ surname-mismatched queries, the open-accuracy gains are $28.51\pm1.16$ points in the uniform pool and $22.22\pm0.67$ points in the lexical pool.

In this preregistered comparison, known-direction labels use mean-token scores and reverse labels use summed-token DC scores. We average main-template scores across the four teachers for both label sets. Label selection does not use the queries' gold answers. We train twelve students using the primary MDM-0.6B recipe across two pools, two label sources, and three seeds, with one RTX A6000 GPU for each run. The trainer withholds the forward edges of all $535$ accepted children of the $512$ query parents and keeps $9{,}757$ forward edges, making every trained query exposure-free.

\begin{table}[htbp]
\caption{MDM-0.6B student results by training run on the unscreened cohort ($512$ trained queries in each pool). Count pairs give known-direction/summed-DC students. Inv.\ counts outputs whose inventory-matched name equals the selected label. Open gains carry paired bootstrap intervals and descriptive McNemar tests.}
\label{tab:prospectivestudents}
\centering\small
\setlength{\tabcolsep}{3pt}
\begin{tabular*}{\linewidth}{@{\extracolsep{\fill}}llrrrrlrr@{}}
\toprule
\tblhead{Pool} & \tblhead{Run} & \tblhead{Open} & \tblhead{Whole} & \tblhead{Matched} & \tblhead{Inv.} & \tblhead{Open gain [95\% CI]} & \tblhead{W/L} & \tblhead{$p$}\\
\midrule
Uniform & A & 440/365 & 440/365 & 448/375 & 509/510 & 14.65 $[11.13,18.16]$ & 84/9 & $2.2\times10^{-16}$\\
 & B & 435/368 & 435/368 & 449/375 & 510/511 & 13.09 $[9.38,16.80]$ & 85/18 & $1.4\times10^{-11}$\\
 & C & 435/371 & 434/371 & 447/375 & 509/511 & 12.50 $[8.98,16.21]$ & 79/15 & $1.2\times10^{-11}$\\
\addlinespace[2pt]
Lexical & A & 359/275 & 359/275 & 365/285 & 512/509 & 16.41 $[12.30,20.70]$ & 110/26 & $1.7\times10^{-13}$\\
 & B & 359/282 & 359/282 & 364/286 & 511/510 & 15.04 $[11.13,19.14]$ & 100/23 & $1.2\times10^{-12}$\\
 & C & 355/280 & 354/280 & 363/286 & 510/512 & 14.65 $[10.55,18.95]$ & 103/28 & $2.8\times10^{-11}$\\
\bottomrule
\end{tabular*}
\end{table}

\FloatBarrier
\section{Stronger comparators and reversed queries}
\label{app:comparators}
\paragraph{Preregistration and scoring.} Table~\ref{tab:comparatorvalues} lists the label accuracies plotted in Figure~\ref{fig:comparators}a, and Table~\ref{tab:flip} gives the accuracies used to compute the differences in Figure~\ref{fig:comparators}b. These comparisons use preregistered analysis plans. Teachers score in BF16 with raw completion prompts, one space between prompt and continuation, and an exact token-prefix check on every scored pair. Acquired-pool intervals resample parents in clusters, and unscreened intervals resample the $512$ parents, each with $10{,}000$ draws.

\begin{table}[htbp]
\caption{Label accuracy (\%) of every comparator on the acquired pools and the two unscreened candidate lists. Blue marks known-direction labels.}
\label{tab:comparatorvalues}
\centering\small
\begin{tabular}{lrrr}
\toprule
\tblhead{Label source} & \tblhead{Acquired, $1{,}390$} & \tblhead{Uniform, $512$} & \tblhead{Lexical, $512$}\\
\midrule
\rowcolor{oursrow}[0pt][0pt]
\textbf{Known direction} & 90.72 & 87.89 & 71.29\\
Tuned reverse, own $\lambda$ & 81.80 & 76.37 & 58.79\\
Tuned reverse, transferred & 80.65 & 74.61 & 57.23\\
DC reverse, summed & 75.18 & 73.44 & 55.86\\
Surname rule & 72.73 & 56.05 & 45.12\\
Reverse, summed & 71.80 & 60.16 & 46.88\\
Trigram rule & 65.47 & 63.48 & 33.20\\
Inverted generation & 25.76 & 16.99 & 16.80\\
Generation, best rule & 17.05 & 4.10 & 2.73\\
\bottomrule
\end{tabular}
\end{table}

\paragraph{Tuned reverse scores.} Table~\ref{tab:tunedreverse} reports label accuracy for four tuned reverse scoring rules, a comparator selected on other cohorts, and known-direction scoring. In all three cohorts, known-direction labels are more accurate even when each reverse scoring rule uses its best coefficient on that cohort. To estimate candidate-name priors without specifying the query parent, we also use the prompt \texttt{Someone is the parent of} on the unscreened cohort. With this prior correction, summed-token reverse scores reach $75.59\%$ and $57.81\%$ on the uniform and lexical pools, respectively, both with $\lambda=0.75$.

Each family averages the corrected teacher scores before selecting the highest-scoring candidate. For teacher $r$, $\ell_r^{\Sigma}$ and $\ell_r$ denote summed and mean continuation log probabilities. The background families score $\ell_r^{\Sigma}(c\mid t_{\mathrm{child}}(p_i))-\lambda\,\ell_r^{\Sigma}(c\mid\texttt{The child is})$ and its mean-token analogue. The Monte Carlo families replace the background with
\begin{equation}
B_r(c)=\log\!\left[\frac{1}{|S_c|}\sum_{p'\in S_c}\exp \ell_r^{\Sigma}\!\left(c\mid t_{\mathrm{child}}(p')\right)\right].
\label{eq:mc}
\end{equation}
The mean-token form divides $B_r(c)$ by teacher $r$'s mean continuation-token count over $S_c$. We sample $32$ reference parents once from Wikidata, excluding screened training and held-out query parents and the unscreened query parents. For each candidate $c$, $S_c$ removes its recorded parents from this shared set.

The grid is $\lambda\in\{0,0.05,\ldots,1.5\}$, and ties prefer the value closest to one, then the smaller value. The acquired pools take the form and $\lambda$ that maximize mean accuracy over the two unscreened pools, and each unscreened pool takes those that maximize acquired-pool accuracy. At $\lambda=0$ and $\lambda=1$, the background families reproduce the reverse and DC labels exactly. Exact McNemar tests receive Holm correction across the three cohorts.

Table~\ref{tab:namecuevalues}a compares label and student accuracy for the known direction and summed DC on queries whose parent and recorded child have different surnames. Known-direction labels and the students trained on them are more accurate in all three cohorts. Extending the label comparison to the transferred tuned-reverse comparator, we find an accuracy advantage of about $22$ to $27$ percentage points for known-direction labels. The paired $95\%$ confidence intervals for their gains over both reverse comparators exclude zero in every cohort. Student gains over summed DC hold in every training run, with similar cohort-mean gains under open and whole-answer evaluation. Known-direction students also have higher mean open accuracy than transferred-label students in both unscreened pools. These gains therefore do not require the parent and recorded child to share a surname.

Table~\ref{tab:strata} compares known-direction, reverse, and DC label accuracy by name-prior stratum. In the highest name-prior tercile, our additional comparison shows that the transferred tuned-reverse comparator improves on DC with coefficient one. Known-direction labels remain more accurate in all three cohorts, even when the reverse correction is selected on other cohorts.

\begin{table}[htbp]
\caption{Tuned reverse label accuracy (\%) with each cohort's own best $\lambda$ in parentheses, and the transferred comparators whose form and $\lambda$ come from the other cohorts. The last two rows give the preregistered contrasts in points with $95\%$ intervals. Blue marks known-direction labels.}
\label{tab:tunedreverse}
\centering\small
\begin{tabular*}{\linewidth}{@{\extracolsep{\fill}}lrrr@{}}
\toprule
\tblhead{Reverse score} & \tblhead{Acquired, $1{,}390$} & \tblhead{Uniform, $512$} & \tblhead{Lexical, $512$}\\
\midrule
Background, summed & 81.58 (0.60) & 76.37 (0.80) & 58.79 (0.75)\\
Background, mean & 80.94 (0.45) & 74.80 (0.70) & 56.84 (0.55)\\
Monte Carlo, summed & 81.80 (0.75) & 76.37 (0.95) & 58.01 (0.90)\\
Monte Carlo, mean & 81.08 (0.60) & 75.98 (0.95) & 55.86 (0.75)\\
\addlinespace[1.5pt]
Transferred & 80.65 (0.75) & 74.61 (0.75) & 57.23 (0.75)\\
\tblindent{form} & background, summed & Monte Carlo, summed & Monte Carlo, summed\\
\rowcolor{oursrow}[0pt][0pt]
\textbf{Known direction} & 90.72 & 87.89 & 71.29\\
Known $-$ transferred & 10.07 $[8.12,12.03]$ & 13.28 $[9.96,16.60]$ & 14.06 $[10.16,17.97]$\\
Wins/losses, Holm $p$ & 170/30, $1.9\times10^{-24}$ & 77/9, $2.7\times10^{-14}$ & 92/20, $3.2\times10^{-12}$\\
\bottomrule
\end{tabular*}
\end{table}

\paragraph{Teacher generation.} Table~\ref{tab:generation} reports each teacher's greedy open accuracy and the label accuracy of its greedy completions and eight-sample majorities mapped onto candidate lists. Even with candidate-list mapping, label accuracy is at most $17.05\%$. In additional comparisons, full-inventory mapping reaches the same maximum accuracy as open matching. The four-teacher majority of greedy completions also trails the best individual teacher's greedy labels after candidate-list mapping in every cohort. These results show that the tested mapping and voting rules leave most reverse labels incorrect.

Each teacher completes \texttt{\{parent\}'s child is} for the $1{,}467$ distinct acquired parents and the $512$ unscreened parents, with at most $16$ new tokens, greedily and with eight samples at temperature $0.7$ and top-$p$ $0.95$. For label construction, we extract a name span by stripping leading quotes and brackets, truncating at the first comma, semicolon, parenthesis, quote, \texttt{who}, \texttt{and}, \texttt{was}, \texttt{is}, or \texttt{born}, and keeping at most four tokens. The $18$ label rules map each teacher's greedy completion or eight-sample majority, and the four-teacher majority of greedy completions, onto either the query's candidate list or the full inventory using the inventory-matching postprocessor of Appendix~\ref{app:implementation}. Sample majorities break ties with the greedy label, and the four-teacher majority breaks ties in teacher order.

\begin{table}[htbp]
\caption{Teacher generation as a label source (\%). Open accuracy credits a true child's full name in the first line of the greedy completion, and pool-mapped columns map the greedy completion or the eight-sample majority onto each candidate list.}
\label{tab:generation}
\centering\small
\setlength{\tabcolsep}{3pt}
\begin{tabular*}{\linewidth}{@{\extracolsep{\fill}}lrrrrrrrr@{}}
\toprule
 & \multicolumn{2}{c}{\tblhead{Open, greedy}} & \multicolumn{3}{c}{\tblhead{Greedy, pool-mapped}} & \multicolumn{3}{c}{\tblhead{Eight samples, pool-mapped}}\\
\cmidrule(lr){2-3}\cmidrule(lr){4-6}\cmidrule(l){7-9}
\tblhead{Teacher} & \tblhead{Acq.} & \tblhead{Unscr.} & \tblhead{Acq.} & \tblhead{Unif.} & \tblhead{Lex.} & \tblhead{Acq.} & \tblhead{Unif.} & \tblhead{Lex.}\\
\midrule
Qwen3-8B & 0.79 & 0.59 & 6.40 & 1.37 & 2.15 & 6.69 & 2.34 & 2.54\\
OLMo-2-7B & 2.37 & 0.98 & 7.41 & 2.54 & 1.56 & 17.05 & 4.10 & 2.73\\
Mistral-7B & 0.79 & 0.39 & 5.90 & 0.98 & 0.98 & 7.91 & 3.12 & 2.73\\
Llama-3.1-8B & 1.65 & 1.17 & 6.55 & 1.95 & 0.98 & 8.42 & 2.54 & 2.73\\
\bottomrule
\end{tabular*}
\end{table}

\paragraph{Inverted generation.} Table~\ref{tab:inversion} reports inverted-generation accuracy, coverage, and precision alongside known-direction accuracy. Most inverted-generation labels are correct, but fewer than $30\%$ of queries receive a label in every cohort. This low coverage leaves its accuracy well below that of known-direction scoring. Additional comparisons show that combining greedy and sampled completions from all four teachers yields higher accuracy than greedy-only voting or the best single teacher, Llama-3.1-8B. Replacing known-direction labels with inverted labels wherever votes are available changes accuracy by at most $0.20$ points in each cohort.

In this prespecified comparison, each teacher completes \texttt{\{child\}'s parent is} for every distinct candidate name in the acquired pools and unscreened lists, greedily and with eight samples under the settings above. The four teachers provide $36$ completions for each candidate. Each completion contributes one vote if its first line names the query parent under the open matcher. We select the eligible candidate with the most votes, breaking ties by candidate order. Queries with no votes receive no label and count as incorrect.

\begin{table}[htbp]
\caption{Inverted generation as a label source (\%). Coverage counts queries with at least one vote, and precision is accuracy on them. The last columns compare known-direction with inverted labels in points, with paired $95\%$ intervals, wins/losses, and Holm-corrected McNemar $p$.}
\label{tab:inversion}
\centering\small
\setlength{\tabcolsep}{3pt}
\begin{tabular*}{\linewidth}{@{\extracolsep{\fill}}lrrrrlrr@{}}
\toprule
\tblhead{Cohort} & \tblhead{Known} & \tblhead{Inverted} & \tblhead{Coverage} & \tblhead{Precision} & \tblhead{Known $-$ inverted} & \tblhead{W/L} & \tblhead{Holm $p$}\\
\midrule
Acquired, $1{,}390$ & \cellcolor{oursrow}90.72 & 25.76 & 27.84 & 92.51 & 64.96 $[62.36,67.65]$ & 913/10 & $1.0\times10^{-254}$\\
Uniform, $512$ & \cellcolor{oursrow}87.89 & 16.99 & 17.19 & 98.86 & 70.90 $[66.99,74.80]$ & 363/0 & $2.1\times10^{-109}$\\
Lexical, $512$ & \cellcolor{oursrow}71.29 & 16.80 & 18.95 & 88.66 & 54.49 $[50.00,58.98]$ & 285/6 & $4.1\times10^{-76}$\\
\bottomrule
\end{tabular*}
\end{table}

\paragraph{Teacherless lexical rules.} Table~\ref{tab:lexicalrules} compares known-direction labels with surname and character-trigram rules on all queries and on the surname-mismatched subset. Known-direction labels are more accurate than both rules in all three candidate pools, including on surname-mismatched queries. On the acquired pools overall, the paired gains over the surname and trigram rules are $17.99$ $[15.74,20.19]$ and $25.25$ $[22.71,27.82]$ points, respectively.

The surname rule selects the alphabetically first eligible candidate whose normalized final token matches the query parent's. If none matches, it selects the alphabetically first candidate, or the summed-DC choice in the fallback variant. The trigram rule selects the candidate with the highest character-trigram Jaccard similarity to the parent name, breaking ties alphabetically.

\begin{table}[htbp]
\caption{Label accuracy (\%) of teacherless lexical rules on all queries and on within-pair surname-mismatched queries ($294$ acquired and $228$ per unscreened pool). Blue marks known-direction labels.}
\label{tab:lexicalrules}
\centering\small
\begin{tabular*}{\linewidth}{@{\extracolsep{\fill}}lrrrrrr@{}}
\toprule
 & \multicolumn{3}{c}{\tblhead{All queries}} & \multicolumn{3}{c}{\tblhead{Surname mismatch}}\\
\cmidrule(lr){2-4}\cmidrule(l){5-7}
\tblhead{Rule} & \tblhead{Acquired} & \tblhead{Uniform} & \tblhead{Lexical} & \tblhead{Acquired} & \tblhead{Uniform} & \tblhead{Lexical}\\
\midrule
\rowcolor{oursrow}[0pt][0pt]
\textbf{Known direction} & 90.72 & 87.89 & 71.29 & 71.43 & 73.25 & 47.81\\
DC reverse, summed & 75.18 & 73.44 & 55.86 & 44.22 & 44.30 & 25.44\\
Surname rule & 72.73 & 56.05 & 45.12 & 6.12 & 1.75 & 2.19\\
Surname rule, DC fallback & 79.14 & 75.00 & 53.91 & 36.39 & 44.30 & 21.93\\
Trigram rule & 65.47 & 63.48 & 33.20 & 37.07 & 24.12 & 7.89\\
\bottomrule
\end{tabular*}
\end{table}

\paragraph{Students on stronger comparator labels.} Table~\ref{tab:comparatorstudents} reports accuracy and label reproduction as three-seed means with sample standard deviations for students trained on known-direction (K), transferred tuned-reverse (T), and generation-majority (G) labels. In paired comparisons, K students exceed T students in open accuracy in every seed, by $12.11$ to $14.65$ points on uniform lists and $13.09$ to $15.04$ points on lexical lists. All six paired confidence intervals exclude zero, with exact McNemar $p\le4.9\times10^{-10}$. K students also exceed G students by at least $68.36$ points in every seed. After inventory matching, mean reproduction of selected labels is $99.5$--$99.9\%$ for all three sources, including the mostly wrong generation labels.

The T and G labels enter the student recipe of Appendix~\ref{app:prospective} with the same withheld forward edges, steps, and seeds. T uses the score form and $\lambda$ selected with gold labels on the other cohorts; G uses the four-teacher majority of greedy completions mapped onto each query's list. Both label sets are selected without using the gold answers of the trained queries.

\begin{table}[htbp]
\caption{Unscreened-cohort MDM-0.6B students trained on known-direction (K), transferred tuned-reverse (T), and generation-majority (G) labels (\%, three-seed mean $\pm$ sample SD). $x$ is label accuracy, and the last columns give label reproduction as a whole string or after inventory matching.}
\label{tab:comparatorstudents}
\centering\small
\setlength{\tabcolsep}{3pt}
\begin{tabular*}{\linewidth}{@{\extracolsep{\fill}}llrrrrrr@{}}
\toprule
 & & & \multicolumn{3}{c}{\tblhead{Accuracy against true children}} & \multicolumn{2}{c}{\tblhead{Reproduces label}}\\
\cmidrule(lr){4-6}\cmidrule(l){7-8}
\tblhead{Pool} & \tblhead{Labels} & \tblhead{$x$} & \tblhead{Open} & \tblhead{Whole} & \tblhead{Matched} & \tblhead{String} & \tblhead{Inventory}\\
\midrule
\rowcolor{oursrow}[0pt][0pt]
Uniform & K & 87.89 & 85.29\sd{0.56} & 85.22\sd{0.63} & 87.50\sd{0.20} & 96.81\sd{0.60} & 99.48\sd{0.11}\\
 & T & 74.61 & 72.20\sd{0.81} & 72.07\sd{0.78} & 74.35\sd{0.11} & 96.81\sd{1.27} & 99.61\sd{0.20}\\
 & G & 0.98 & 0.98\sd{0.00} & 0.98\sd{0.00} & 0.98\sd{0.00} & 98.24\sd{1.09} & 99.61\sd{0.39}\\
\addlinespace[1.5pt]
\rowcolor{oursrow}[0pt][0pt]
Lexical & K & 71.29 & 69.86\sd{0.45} & 69.79\sd{0.56} & 71.09\sd{0.20} & 97.59\sd{0.69} & 99.80\sd{0.20}\\
 & T & 57.23 & 55.66\sd{0.59} & 55.60\sd{0.49} & 57.10\sd{0.23} & 97.59\sd{0.69} & 99.87\sd{0.23}\\
 & G & 1.17 & 1.11\sd{0.11} & 1.11\sd{0.11} & 1.17\sd{0.00} & 98.37\sd{0.30} & 99.80\sd{0.20}\\
\bottomrule
\end{tabular*}
\end{table}

\paragraph{Reversed queries.} Figure~\ref{fig:comparators}b and Table~\ref{tab:flip} compare sentence directions on parent and child queries at $\lambda=1$. On corpus facts, the child-to-parent sentence wins for both query sides and both lists. For child queries, a separate best-$\lambda$ comparison gives gains of $11.52$ points on uniform lists at $\lambda=1.0$ and $6.45$ $[3.12,9.77]$ points on lexical lists at $\lambda=0.9$, with Holm $p\le2.2\times10^{-4}$. At $\lambda=1$, the lexical gain is $5.86$ points. The better sentence direction therefore persists when the query is reversed.

Each unscreened fact becomes a child query whose accepted answers are all recorded parents of the child. Its two $64$-name lists use the sorted parent inventory, exclude the query child, and include the recorded parent. Uniform lists add $63$ pseudorandom distractors. Lexical lists add the $32$ parents with the highest character-trigram similarity to the child and $31$ pseudorandom distractors, with pseudorandom positions. For teacher $r$ and candidate parent $p'$, the direct child-to-parent score is $\ell_r^{\Sigma}(p'\mid\texttt{\{child\}'s parent is})-\lambda\,\ell_r^{\Sigma}(p'\mid\texttt{The parent is})$, and the parent-to-child channel score is the mean-token $\ell_r(c\mid\texttt{\{p'\}'s child is})$. Holm correction covers the two preregistered best-$\lambda$ contrasts.

\paragraph{Facts with a notable parent.} Table~\ref{tab:flip} compares both sentence directions on corpus facts and on newly mined facts with a notable parent. The child-to-parent sentence wins on corpus facts and the parent-to-child sentence on notable-parent facts for both query sides and both lists. Averaging the fact-level child-to-parent minus parent-to-child differences over these four conditions gives $11.82$ $[9.47,14.16]$ points on corpus facts and $-6.86$ $[-8.35,-5.42]$ points on notable-parent facts. On their $228$ and $404$ surname-mismatched facts, the corresponding differences are $21.27$ and $-12.13$ points. The preferred sentence direction tracks which entity is notable.

Mining, candidate lists, scoring, and analysis are preregistered. We mine facts with one Wikidata Query Service query for each of the $19$ corpus occupations. Eligible parents are human, hold a corpus occupation, and have an English Wikipedia article and at least $20$ sitelinks. An eligible child is human, has that parent recorded as father or mother, has no English Wikipedia article and at most $2$ sitelinks, and was born in $1900$ or later. Both names pass corpus cleaning. We exclude facts whose child name is a corpus child name or whose parent is a query parent of either earlier cohort. Pseudorandomly ordered parents contribute one child each until $1{,}024$ facts are selected. Both query sides use uniform and lexical $64$-name lists built from the mined child and parent names as for the unscreened cohort. All four teachers score both sentences and context-only backgrounds with the main template.

\begin{table}[htbp]
\caption{Sentence scoring on corpus and notable-parent facts (\%, four-teacher means, $64$-name lists). C$\to$P conditions on the child and P$\to$C on the parent; the sentence scoring the query entity uses the channel form. W/L counts facts that only one sentence answers.}
\label{tab:flip}
\centering\small
\setlength{\tabcolsep}{2.5pt}
\begin{tabular*}{\linewidth}{@{\extracolsep{\fill}}llrrrrrrrr@{}}
\toprule
 & & \multicolumn{4}{c}{\tblhead{Uniform}} & \multicolumn{4}{c}{\tblhead{Lexical}}\\
\cmidrule(lr){3-6}\cmidrule(l){7-10}
\tblhead{Facts} & \tblhead{Query} & \tblhead{C$\to$P} & \tblhead{P$\to$C} & \tblhead{W/L} & \tblhead{$p$} & \tblhead{C$\to$P} & \tblhead{P$\to$C} & \tblhead{W/L} & \tblhead{$p$}\\
\midrule
Corpus, $512$ & Parent & 87.89 & 73.44 & 82/8 & $1.4\times10^{-16}$ & 71.29 & 55.86 & 101/22 & $2.7\times10^{-13}$\\
 & Child & 86.52 & 75.00 & 67/8 & $1.0\times10^{-12}$ & 73.63 & 67.77 & 52/22 & $6.4\times10^{-4}$\\
\addlinespace[1.5pt]
Notable parent, $1{,}024$ & Parent & 81.35 & 85.06 & 30/68 & $1.6\times10^{-4}$ & 69.43 & 76.27 & 31/101 & $7.6\times10^{-10}$\\
 & Child & 79.98 & 88.18 & 24/108 & $6.4\times10^{-14}$ & 69.63 & 78.32 & 39/128 & $2.9\times10^{-12}$\\
\bottomrule
\end{tabular*}
\end{table}

\section{Default-exposure control recipes}
\label{app:controls}
Table~\ref{tab:studentsizes} extends the default-exposure label-source comparison in Table~\ref{tab:historicalfactual} across student sizes and objectives on all $1{,}500$ trained queries. Rows have three seeds except the AR-8B single-teacher row, which uses five, and AR students use the token budget described in Appendix~\ref{app:implementation}. Across student sizes and objectives, single-teacher, agreement, and gold labels keep the same ordering, and the larger MDM mainly raises open accuracy at fixed label correctness. Table~\ref{tab:historicalvariants} lists the label variants omitted from Table~\ref{tab:historicalfactual}.

\begin{table}[htbp]
\caption{Default-exposure students of two MDM sizes and an AR-8B student on all $1{,}500$ trained queries (\%, mean $\pm$ sample SD). Blue marks eight-channel agreement labels.}
\label{tab:studentsizes}
\centering\small
\setlength{\tabcolsep}{3pt}
\begin{tabular*}{\linewidth}{@{\extracolsep{\fill}}lrrrrrrr@{}}
\toprule
 & & \multicolumn{2}{c}{\tblhead{MDM-4B}} & \multicolumn{2}{c}{\tblhead{MDM-0.6B}} & \multicolumn{2}{c}{\tblhead{AR-8B}}\\
\cmidrule(lr){3-4}\cmidrule(lr){5-6}\cmidrule(l){7-8}
\tblhead{Labels} & \tblhead{$x$} & \tblhead{Open} & \tblhead{Matched} & \tblhead{Open} & \tblhead{Matched} & \tblhead{Open} & \tblhead{Matched}\\
\midrule
Single (Qwen3-8B) & 81.13 & 75.22\sd{0.37} & 80.51\sd{0.20} & 70.67\sd{0.69} & 80.09\sd{0.17} & 78.51\sd{0.25} & 79.23\sd{0.13}\\
\rowcolor{oursrow}[0pt][0pt]
Agreement (8) & 91.53 & 84.33\sd{1.62} & 90.73\sd{0.24} & 80.51\sd{1.11} & 90.38\sd{0.34} & 88.13\sd{0.50} & 89.11\sd{0.47}\\
Gold & 100.00 & 92.00\sd{2.07} & 98.91\sd{0.04} & 83.58\sd{1.35} & 97.40\sd{0.37} & 96.98\sd{0.14} & 97.73\sd{0.07}\\
\bottomrule
\end{tabular*}
\end{table}

\paragraph{Control arms.}
All control arms use the optimizer and LoRA settings specified in Section~\ref{sec:setup}. Warm trains only the forward stage. MLM-U replaces both stages with one $8{,}800$-step stage on the same forward corpus, masking a uniformly sampled number of prompt and answer positions, and uses no reverse labels. Random labels draw one name uniformly from each query's stored top-eight single-teacher candidates with the run seed. Self-ranked labels come from the student's Qwen3-4B base model, which ranks inventory children after \texttt{\{parent\}'s child is} and reranks its top candidates by a contrast against reference parents. Majority keeps a name chosen by at least two of the three acquisition scans and otherwise uses Qwen3-8B. The alternate-template mean averages per-candidate z-scores of the three acquisition teachers under the alternate template. The AR identity-bridge control adapts \citet{ma2026breaking} to our total warm and SFT token budget. One stage mixes six copies of the forward corpus with name-identity sentences and decomposed identity sentences, without reverse labels, under our optimization settings.

\begin{figure}[htbp]
\centering
\includegraphics[width=\linewidth]{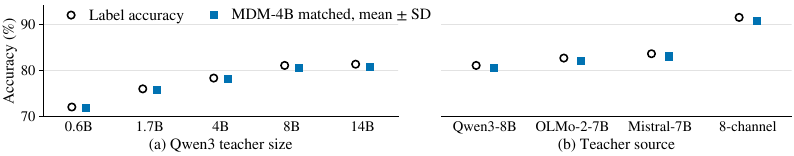}\par
\caption{Teacher scaling and transfer with default forward exposure. Circles show label accuracy and squares show MDM-4B matched accuracy with three-seed sample SD.}
\label{fig:teachers}
\end{figure}

\paragraph{Teacher scaling.} Figure~\ref{fig:teachers} and Table~\ref{tab:teachervalues} compare label accuracy and MDM-4B matched accuracy across teacher sizes and sources under default exposure. Label accuracy rises with Qwen3 teacher size, and matched student accuracy follows it closely. The student's own Qwen3-4B base model also supplies useful known-direction labels, with $78.40\%$ accuracy. Each single-teacher pipeline uses that teacher for candidate acquisition and label scoring, and each Qwen3 teacher in Figure~\ref{fig:teachers}a scans the inventory independently. The pipelines in the right panel also differ in candidate acquisition and self-identity filtering. The self-ranked control instead uses the student's base model for reverse inventory ranking with a reference-parent rerank.

\begin{table}[htbp]
\caption{Single-teacher label accuracy and default-exposure MDM-4B matched accuracy (\%, three-seed mean $\pm$ sample SD) by teacher size and source. Eight channels is the agreement rule over four teachers and two templates.}
\label{tab:teachervalues}
\centering\small
\setlength{\tabcolsep}{3pt}
\begin{minipage}[t]{0.47\linewidth}
\centering
\textbf{(a) Qwen3 teacher size}\par\smallskip
\begin{tabular*}{\linewidth}{@{\extracolsep{\fill}}lrr@{}}
\toprule
\tblhead{Teacher} & \tblhead{Label} & \tblhead{Matched}\\
\midrule
Qwen3-0.6B & 72.13 & 71.96\sd{0.08}\\
Qwen3-1.7B & 76.07 & 75.84\sd{0.10}\\
Qwen3-4B & 78.40 & 78.13\sd{0.20}\\
Qwen3-8B & 81.13 & 80.51\sd{0.20}\\
Qwen3-14B & 81.40 & 80.80\sd{0.07}\\
\bottomrule
\end{tabular*}
\end{minipage}\hfill
\begin{minipage}[t]{0.47\linewidth}
\centering
\textbf{(b) Teacher source}\par\smallskip
\begin{tabular*}{\linewidth}{@{\extracolsep{\fill}}lrr@{}}
\toprule
\tblhead{Teacher} & \tblhead{Label} & \tblhead{Matched}\\
\midrule
Qwen3-8B & 81.13 & 80.51\sd{0.20}\\
OLMo-2-7B & 82.73 & 82.16\sd{0.23}\\
Mistral-7B & 83.67 & 83.07\sd{0.24}\\
Eight channels & 91.53 & 90.73\sd{0.24}\\
\bottomrule
\end{tabular*}
\end{minipage}
\end{table}

\paragraph{Name-cue and decoding controls.}
Table~\ref{tab:namecuevalues} reports accuracy for the surname-mismatch and inventory-enlargement controls plotted in Figure~\ref{fig:namecues}, and Table~\ref{tab:namecontrolsfull} adds student results on the surname strata and under three decoding orders. The default-exposure surname stratum contains $305$ trained queries with different parent and child final name tokens. The inventory test adds $9{,}498$ parent-only names to the original $8{,}218$ and reuses every generation, and surname-only lookup breaks ties with the first sorted candidate. The decoding test retrains three single-teacher-label checkpoints and decodes each under all three orders, so its confidence-order row differs slightly from the single-teacher runs of Table~\ref{tab:historicalfactual}. Decoding order moves open accuracy by up to $7.5$ points but matched accuracy by at most $0.2$ points.

\begin{table}[htbp]
\caption{Name-cue controls (\%, three-seed mean $\pm$ sample SD). (a) Label accuracy and MDM-0.6B student open accuracy on surname-mismatched queries. (b) Default-exposure MDM-4B matched accuracy with the original and enlarged inventories. Blue marks known-direction labels.}
\label{tab:namecuevalues}
\centering\small
\setlength{\tabcolsep}{3pt}
\begin{minipage}[t]{0.54\linewidth}
\centering
\textbf{(a) Surname-mismatched queries}\par\smallskip
\begin{tabular*}{\linewidth}{@{\extracolsep{\fill}}lrrrr@{}}
\toprule
 & \multicolumn{2}{c}{\tblhead{Labels}} & \multicolumn{2}{c}{\tblhead{Students}}\\
\cmidrule(lr){2-3}\cmidrule(l){4-5}
\tblhead{Cohort} & \tblhead{Known} & \tblhead{DC} & \tblhead{Known} & \tblhead{DC}\\
\midrule
Acquired, $294$ & \cellcolor{oursrow}71.43 & 44.22 & \cellcolor{oursrow}53.17\sd{3.76} & 30.95\sd{1.48}\\
Uniform, $228$ & \cellcolor{oursrow}73.25 & 44.30 & \cellcolor{oursrow}70.32\sd{0.91} & 41.81\sd{0.51}\\
Lexical, $228$ & \cellcolor{oursrow}47.81 & 25.44 & \cellcolor{oursrow}46.05\sd{0.44} & 23.83\sd{0.67}\\
\bottomrule
\end{tabular*}
\end{minipage}\hfill
\begin{minipage}[t]{0.42\linewidth}
\centering
\textbf{(b) Inventory enlargement}\par\smallskip
\begin{tabular*}{\linewidth}{@{\extracolsep{\fill}}lrr@{}}
\toprule
\tblhead{Labels} & \tblhead{Original} & \tblhead{Enlarged}\\
\midrule
Gold & 98.91\sd{0.04} & 98.16\sd{0.17}\\
\rowcolor{oursrow}[0pt][0pt]
Agreement (8) & 90.73\sd{0.24} & 89.98\sd{0.60}\\
Single & 80.51\sd{0.20} & 79.73\sd{0.31}\\
Surname-only & 69.80 & 35.67\\
Warm start & 39.56\sd{3.00} & 10.69\sd{1.17}\\
\bottomrule
\end{tabular*}
\end{minipage}
\end{table}

\begin{table}[htbp]
\caption{Default-exposure MDM-4B students trained on single-teacher and combined known-direction labels (\%, mean $\pm$ sample SD over three seeds). Agreement rows differ only in the number of channels.}
\label{tab:historicalvariants}
\centering\small
\begin{tabular}{lrrr}
\toprule
\tblhead{Labels} & \tblhead{$x$} & \tblhead{Open} & \tblhead{Matched}\\
\midrule
Single teacher & 81.13 & $75.22\pm0.37$ & $80.51\pm0.20$\\
Majority of three scans & 84.60 & $79.47\pm0.20$ & $83.98\pm0.17$\\
Alternate-template mean (3) & 90.33 & $83.36\pm1.60$ & $89.56\pm0.23$\\
\addlinespace[1.5pt]
Agreement, 6 channels & 90.20 & $84.24\pm0.83$ & $89.44\pm0.17$\\
\rowcolor{oursrow}[0pt][0pt]
Agreement, 8 channels & 91.53 & $84.33\pm1.62$ & $90.73\pm0.24$\\
Agreement, 10 channels & 91.07 & $84.60\pm0.81$ & $90.27\pm0.07$\\
Agreement, 12 channels & 90.80 & $83.33\pm1.75$ & $89.80\pm0.13$\\
\bottomrule
\end{tabular}
\end{table}

\begin{table}[htbp]
\caption{Name-cue and decoding controls (\%, three-seed mean $\pm$ sample SD). (a) Primary MDM-0.6B label sources on $294$ surname-mismatched exposure-free queries. (b) Default-exposure MDM-4B students on $305$ surname-mismatched queries. (c) Retrained MDM-4B single-teacher checkpoints under three decoding orders.}
\label{tab:namecontrolsfull}
\centering\small
\begin{tabular}{lrr}
\toprule
\tblhead{Labels or order} & \tblhead{Open} & \tblhead{Matched}\\
\midrule
\rowcolor{groupbg}[0pt][0pt]\multicolumn{3}{@{}c@{}}{\textbf{(a) Primary MDM-0.6B, $n=294$}\strut}\\
\rowcolor{oursrow}[0pt][0pt]
Known mean & $53.17\pm3.76$ & $68.82\pm0.39$\\
Reverse mean & $33.22\pm1.09$ & $39.68\pm1.09$\\
DC reverse mean & $21.20\pm0.20$ & $24.04\pm0.52$\\
Llama two-template & $52.83\pm1.04$ & $71.43\pm0.68$\\
Agreement (8) & $55.56\pm1.37$ & $71.43\pm0.59$\\
\midrule
\rowcolor{groupbg}[0pt][0pt]\multicolumn{3}{@{}c@{}}{\textbf{(b) Default-exposure MDM-4B, $n=305$}\strut}\\
Warm & $2.95\pm0.66$ & $17.60\pm1.33$\\
Single & $49.18\pm0.87$ & $55.19\pm0.19$\\
\rowcolor{oursrow}[0pt][0pt]
Agreement (8) & $64.37\pm2.42$ & $73.55\pm0.50$\\
Gold & $86.99\pm5.12$ & $97.70\pm0.57$\\
\midrule
\rowcolor{groupbg}[0pt][0pt]\multicolumn{3}{@{}c@{}}{\textbf{(c) Retrained MDM-4B checkpoints by decoding order, $n=1{,}500$}\strut}\\
Confidence & $76.04\pm1.28$ & $80.73\pm0.07$\\
Left-to-right & $77.40\pm0.52$ & $80.73\pm0.18$\\
Random & $69.89\pm1.46$ & $80.53\pm0.44$\\
\bottomrule
\end{tabular}
\end{table}

\begin{figure}[htbp]
\centering
\includegraphics[width=\linewidth]{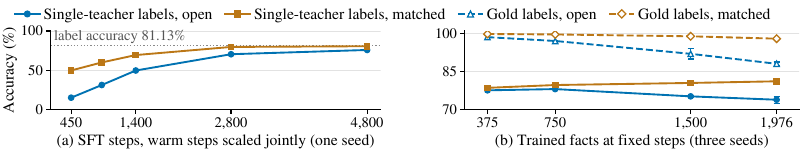}\par
\caption{Optimization controls with default forward exposure. (a) Warm and SFT budgets vary jointly for single-teacher labels. (b) The number of trained facts varies at fixed steps (mean $\pm$ sample SD). The panels use different accuracy ranges.}
\label{fig:optimization}
\end{figure}

\paragraph{Training budget and number of facts.}
\label{app:budget}
Figure~\ref{fig:optimization} compares open and inventory-matched student accuracy as warm and SFT budgets vary jointly and as the number of trained facts varies at fixed steps. Table~\ref{tab:budget} lists the exact values, label accuracies, and expected presentations of each fact. Longer training raises both accuracy metrics and brings matched accuracy closer to label accuracy. At fixed steps, open and matched accuracy decline as the number of gold-labeled facts increases. With single-teacher labels, matched accuracy remains close to label accuracy as both rise across the nested query sets.

The joint-budget ladder uses single-teacher labels with accuracy $81.13\%$, one seed, forward mixing, and $1{,}500$ trained queries. The fact-count ladder fixes $4{,}000/4{,}800$ steps and varies nested reverse-query prefixes, changing label accuracy and repetitions of each fact. Each size evaluates its own trained queries over three seeds.

\begin{table}[!htb]
\caption{Optimization and fact-count controls with default forward exposure (accuracy in percent). The joint-budget panel uses one seed; the fixed-step panel reports three-seed mean $\pm$ sample SD. Views count expected reverse-example presentations.}
\label{tab:budget}
\centering\small
\textbf{Joint warm and SFT budget, single-teacher labels}\par\smallskip
\begin{tabularx}{\linewidth}{*{5}{>{\raggedleft\arraybackslash}X}}
\toprule
\tblhead{Warm steps} & \tblhead{SFT steps} & \tblhead{Open} & \tblhead{Matched} & \tblhead{Matched / $x$ (\%)}\\
\midrule
400 & 450 & 15.13 & 49.80 & 61.38\\
800 & 900 & 31.27 & 60.00 & 73.95\\
1,200 & 1,400 & 49.73 & 69.40 & 85.54\\
2,400 & 2,800 & 70.47 & 79.73 & 98.27\\
4,000 & 4,800 & 75.87 & 80.67 & 99.42\\
\bottomrule
\end{tabularx}
\par\medskip
\textbf{Number of facts at fixed steps}\par\smallskip
\begin{tabularx}{\linewidth}{rr*{5}{>{\raggedleft\arraybackslash}X}}
\toprule
 &  & \multicolumn{2}{c}{\tblhead{Gold labels}} & \multicolumn{3}{c}{\tblhead{Single-teacher labels}}\\
\cmidrule(lr){3-4}\cmidrule(lr){5-7}
\tblhead{$N$} & \tblhead{Views} & \tblhead{Open} & \tblhead{Matched} & \tblhead{Label $x$} & \tblhead{Open} & \tblhead{Matched}\\
\midrule
375 & 102.4 & $98.58\pm0.62$ & $99.82\pm0.15$ & 78.67 & $77.60\pm0.71$ & $78.58\pm0.15$\\
750 & 51.2 & $97.07\pm0.13$ & $99.64\pm0.15$ & 79.73 & $78.09\pm0.20$ & $79.69\pm0.08$\\
1,500 & 25.6 & $92.00\pm2.07$ & $98.91\pm0.04$ & 81.13 & $75.22\pm0.37$ & $80.51\pm0.20$\\
1,976 & 18.8 & $88.07\pm0.71$ & $97.98\pm0.45$ & 82.14 & $73.89\pm1.39$ & $81.16\pm0.34$\\
\bottomrule
\end{tabularx}
\end{table}

\paragraph{Selective supervision.}
Table~\ref{tab:selectivepaired} compares selected-label accuracy, supervision coverage, and student matched accuracy for margin filtering, unanimity, and full-coverage agreement. Filtering labels for precision leaves many queries without reverse supervision, so full-coverage agreement gives the highest matched accuracy over all trained queries. Margin filtering keeps Qwen3-8B labels when the two highest known-direction scores in the full inventory differ by at least $0.905$, a threshold calibrated with reference answers. Unanimity keeps a label only when Qwen3-8B, OLMo-2-7B, and Mistral-7B-v0.3 select the same full-inventory maximizer.

\begin{table}[htbp]
\caption{Selective supervision for default-exposure MDM-4B students on retained and all trained queries (accuracies in \%, mean $\pm$ sample SD over three seeds). $x$ is selected-label accuracy. Blue marks the full-coverage known-direction rule.}
\label{tab:selectivepaired}
\centering\small
\begin{tabularx}{\linewidth}{>{\raggedright\arraybackslash}Xrrrr}
\toprule
\tblhead{Rule} & \tblhead{Kept (queries)} & \tblhead{$x$} & \tblhead{Matched kept} & \tblhead{Matched all $1{,}500$}\\
\midrule
Margin & 827 & 98.19 & 98.02\sd{0.18} & 70.98\sd{1.24}\\
Unanimity & 1,071 & 96.08 & 95.74\sd{0.35} & 77.44\sd{0.04}\\
\rowcolor{oursrow}[0pt][0pt]
\textbf{Eight-channel agreement} & 1,500 & 91.53 & 90.73\sd{0.24} & 90.73\sd{0.24}\\
\bottomrule
\end{tabularx}
\end{table}

\end{document}